%% file: tacl2021v1-template.tex
\documentclass[11pt,a4paper]{article}
\usepackage{lineno}
\usepackage{times,latexsym}
\usepackage{url}
\usepackage[T1]{fontenc}

\usepackage{times}
\usepackage{latexsym}
\usepackage{graphicx}
\usepackage{xcolor}
\usepackage[skip=2pt]{caption}
\usepackage[hang,flushmargin]{footmisc}
\usepackage{amssymb}
\usepackage{booktabs}
\usepackage{multirow}
\usepackage{xcolor}
\usepackage{tcolorbox}
\usepackage{lipsum}
\usepackage{amsmath}
\usepackage{tikz}
\usepackage{subcaption}
\usepackage{soul}
\usepackage[normalem]{ulem}
\usepackage[table]{xcolor}
\usepackage{colortbl}
\usepackage{hhline}
\usepackage{makecell}
\usepackage{multirow}
\usepackage{tabularx}
\usepackage[linesnumbered, ruled, vlined]{algorithm2e}

\definecolor{color1}{rgb}{1.0, 0.97, 0.85} 
\definecolor{color2}{rgb}{1.0, 0.92, 0.65}
\definecolor{color3}{rgb}{1.0, 0.85, 0.45}
\definecolor{color4}{rgb}{1.0, 0.75, 0.30}
\definecolor{color5}{rgb}{0.95, 0.60, 0.20}

\usepackage[acceptedWithA]{tacl2021v1}

\usepackage{xspace,mfirstuc,tabulary}

\newif\iftaclinstructions
\taclinstructionsfalse 
\iftaclinstructions
\renewcommand{\confidential}{}
\renewcommand{\anonsubtext}{(No author info supplied here, for consistency with
TACL-submission anonymization requirements)}
\newcommand{\instr}
\fi

\iftaclpubformat 

\else

\fi

\title{\textsc{DyCP:} Dynamic Context Pruning for Long-Form Dialogue with LLMs}

\author{
  Nayoung Choi
  \\
  Computer Science
  \\
  Emory University
  \\
  Atlanta, Georgia, USA
  \\
  \texttt{\small{nayoung.choi@emory.edu}}
  \And
  Jonathan Zhang
  \\
  Computer Science
  \\
  Emory University
  \\
  Atlanta, Georgia, USA
  \\
  \texttt{\small{jonathan.zhang2@emory.edu}}
  \And
  Jinho D. Choi
  \\
  Computer Science
  \\
  Emory University
  \\
  Atlanta, Georgia, USA
  \\
  \texttt{\small{jinho.choi@emory.edu}}
}

\date{}

\begin{document}
\maketitle

\begin{abstract}
\input{latex/section/abstract}
\end{abstract}

\section{Introduction}
\input{latex/section/introduction}

\section{Related Work}
\label{sec:related_work}
\input{latex/section/related_work}

\section{Method}
\label{sec:method}
\input{latex/section/method}

\section{Experiment}
\label{sec:experiment}
\input{latex/section/experiment}

\section{Analysis}
\label{sec:analysis}
\input{latex/section/analysis}

\section{Discussion}
\input{latex/section/discussion}

\section{Conclusion}
\input{latex/section/conclusion}

\section{Limitations and Practical Considerations}
\input{latex/section/limitation}

\bibliography{tacl2021}
\bibliographystyle{acl_natbib}

\clearpage
\appendix
\section{Appendix}
\input{latex/section/appendix}

\end{document}

%% file: latex/section/abstract.tex
Large Language Models (LLMs) increasingly operate over long-form dialogues with frequent topic shifts. While recent LLMs support extended context windows, efficient management of dialogue history in practice is needed due to inference cost and latency constraints. We present \textsc{DyCP}, a lightweight context management method implemented outside the LLM that dynamically identifies and retrieves relevant dialogue segments conditioned on the current turn, without offline memory construction. \textsc{DyCP} manages dialogue context while preserving the sequential nature of dialogue without predefined topic boundaries, enabling adaptive and efficient context selection. Across three long-form dialogue benchmarks—\texttt{LoCoMo}, \texttt{MT-Bench+}, and \texttt{SCM4LLMs}—and multiple LLM backends, \textsc{DyCP} achieves competitive answer quality in downstream generation, with more selective context usage and improved inference efficiency.

%% file: latex/section/introduction.tex
Large Language Models (LLMs)-powered dialogue systems support long-form, open-domain conversations that span hundreds of turns, where users frequently shift topics and revisit earlier subjects over the course of interaction. To preserve contextual coherence, many systems prepend the entire dialogue history to each new user utterance. While this approach helps maintain dialogue continuity, it introduces growing overhead. Longer contexts increase inference costs—whether in GPU resources or API usage—and slow down response latency. As the context expands, the model also faces a \textit{needle-in-a-haystack} problem \cite{kamradt2023needle}, where identifying relevant information (the \textit{needle}) becomes difficult, as key details are often buried in the surrounding dialogue (the \textit{haystack}).

To mitigate these issues, alongside ongoing efforts to improve LLMs’ intrinsic long-context handling capabilities, prior work has explored external context management strategies, including summarization~\citep{wang2025recursivelysummarizingenableslongterm, langchain_summary_memory_2023} and retrieval of past turns, either at the turn level~\citep{maharana2024evaluatinglongtermconversationalmemory} or at the segment level~\citep{pan2025memoryconstructionretrievalpersonalized, lu2023memochattuningllmsuse}. However, these external approaches involve trade-offs that introduce practical challenges. Summarization may omit key details necessary for the current turn. Although turn-level retrieval is query-targeted and efficient, it can disrupt coherence by ignoring discourse continuity and dependencies. Segment-level retrieval better preserves context by grouping related turns into topic-level segments, but existing methods assume fixed topic boundaries defined based on past dialogue context through pre-segmentation. In practice, however, these boundaries may vary with the current user utterance, suggesting that segmentation should instead be query-dependent. In addition, pre-segmentation commonly depends on extra LLM calls, which can increase overall cost.

\begin{figure*}[htp!]
    \centering
    \includegraphics[width=1.0\linewidth]{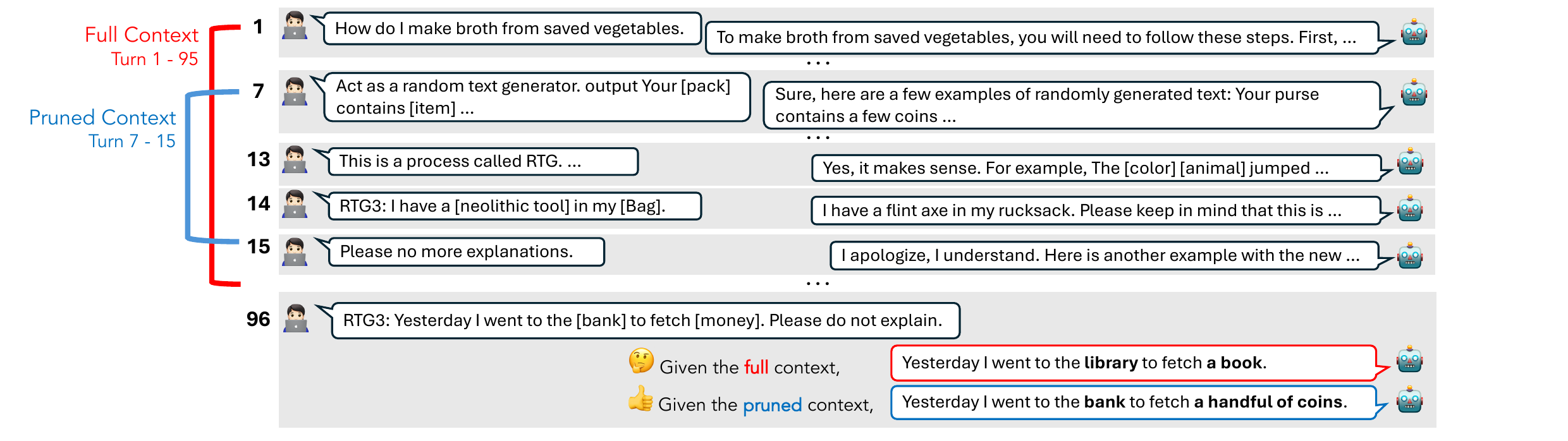}
    \caption{For the current user utterance {\small (Turn 96)}, which requests the RTG3 task defined earlier in the conversation, the agent generates terms like \textit{bank} and \textit{a handful of coins} when using \textcolor{blue}{the pruned context} selected by our method {\small (Turns 7–15)}, accurately reflecting the task intent. In contrast, using \textcolor{red}{the full context} {\small (Turns 1–95)} results in outputs like \textit{library} and \textit{book}, which are plausible but slightly off-topic for the [bank] and [money] theme. While this example uses top-1 pruning, our method generally selects multiple segments dynamically. Pruning also reduces first-token latency about three times in this example.}
    \label{fig:example}
\vspace{-0.3cm}
\end{figure*}

This paper introduces \textsc{DyCP}, a dynamic context pruning method that identifies and retrieves relevant dialogue segments at query time without relying on pre-segmentation. Given a current user query, \textsc{DyCP} estimates its relevance to previous dialogue turns and identifies a sequence of consecutive spans with sustained relevance using \texttt{KadaneDial}, our extension of Kadane’s algorithm\footnote{The original Kadane’s algorithm\cite{10.1145/358234.381162} is a classic linear-time method for finding the maximum subarray.}. \textsc{DyCP} preserves the dialogue context relevant to the current turn while substantially reducing the amount of preceding context included in the input. Figure~\ref{fig:example} illustrates this behavior with an example from a 96-turn conversation in the SCM4LLMs dataset~\cite{wang2024enhancing}, using GPT-4o\footnote{\url{https://platform.openai.com/docs/models/gpt-4o}} as the backend model. In this example, conditioning on only 9 turns of pruned context leads the model to produce a response that is more closely aligned with the user intent than one generated using the full dialogue history, while operating on a substantially shorter input.

In this work, we examine how \textsc{DyCP} differs from existing context management methods through empirical analysis and discuss the relationship between modern LLMs’ expanded context windows, their practical long-context processing capacity, and the role of context management methods such as \textsc{DyCP}. We conduct experiments on three long-form dialogue benchmarks and compare \textsc{DyCP} with six alternative approaches across a range of closed- and open-source LLM backends that support chat-style interaction and long context windows of at least 128k tokens. In summary, our contributions are as follows:
\vspace{-0.2cm}
\begin{itemize}
\setlength\itemsep{-0.05cm}
\item We introduce \textsc{DyCP}, a lightweight context management method that dynamically selects coherent and query-relevant dialogue segments without pre-segmentation or additional LLM calls (Section~\ref{sec:method}).
\item We provide a systematic empirical comparison of \textsc{DyCP} against six other approaches across three benchmarks and a range of closed- and open-source LLM backends, highlighting when and where dynamic context pruning is effective (Section~\ref{sec:experiment}).
\item We offer an analysis of long-context behavior in modern LLMs, examining how dialogue length, the position of relevant context, and retrieval recall interact to shape context management effectiveness, and distilling practical insights for method design (Section~\ref{sec:analysis}).
\end{itemize}

%% file: latex/section/related_work.tex
\noindent \textbf{Long-Context Capabilities of LLMs} $\:$ While recent LLMs increasingly support longer context windows as model scale grows, long-context processing has also benefited from advances in fine-tuning and inference techniques. \citet{chen2023extendingcontextwindowlarge} propose a fine-tuning method that extends the model's positional encoding to support longer contexts while preserving original performance. Building on this, \citet{peng2024yarn} enhance fine-tuning efficiency, enabling models to handle up to 128k tokens with significantly less training. Furthermore, \citet{10.5555/3692070.3692130} introduce an inference-time approach that chunks the input text and links the segments, allowing long-context processing without any additional training.

\vspace{0.3cm}
\noindent \textbf{External Context Management Methods} $\:$ Alongside advances in model-side long-context capabilities, a line of work has focused on managing dialogue history outside the model by reducing the amount of context referenced at inference time. To evaluate context management strategies in long-form open-domain conversations, \citet{maharana2024evaluatinglongtermconversationalmemory} introduces the \texttt{LoCoMo} benchmark and shows that turn-level history retrieval produces better responses than feeding the entire dialogue history to LLMs. \citet{lu2023memochattuningllmsuse} presents the \texttt{MT-Bench+} benchmark and proposes \texttt{MemoChat}, a segment-level retrieval approach that leverages LLMs for both segmentation and retrieval. \citet{pan2025memoryconstructionretrievalpersonalized} demonstrates that segment-level retrieval outperforms turn-level and session-level methods, and presents \texttt{SeCom}, which combines LLM-based segmentation, lightweight retrieval, and context denoising. \citet{wang2024enhancing} introduces the \texttt{SCM4LLMs} benchmark and proposes a turn-level retrieval mechanism composed of multiple LLM-based agents, including memory summarization and memory controller modules. In addition, \citet{yuan-etal-2025-personalized} present \texttt{CondMem}, a hybrid method that combines summarization-based dialogue compression with selective memory construction.


%% file: latex/section/method.tex
This section describes how \textsc{DyCP} dynamically manages dialogue context in long-form conversations. Given the current user query at turn $n$, denoted as $q_n$, the dialogue history is represented as $\mathbf{H} = [h_1, h_2, \ldots, h_{n-1}]$, where each element corresponds to one of the preceding $n{-}1$ turns. Each turn $h_k$ consists of a user–agent utterance pair $(q_k, a_k)$. Dialogue turns are processed independently as they occur: each utterance pair is concatenated into a single textual unit, $h_k = [q_k; a_k]$, and encoded once using a bi-encoder retriever $\texttt{B}$.\footnote{A bi-encoder independently encodes each text input using shared parameters, enabling efficient embedding-based similarity computation~\cite{lei-etal-2023-unsupervised}.} The resulting turn-level embeddings are stored incrementally, forming a sequence of embeddings available at turn $n$:
\[
\begin{split}
\mathbf{H} = [h_1, h_2, \ldots, h_{n{-}1}]; \;\; h_k=[q_k; a_k] \\
\mathbf{H}^{\text{emb}} = [\texttt{B}(h_1),\texttt{B}(h_2), \ldots, \texttt{B}(h_{n{-}1})] \in \mathbb{R}^{(n-1)\times d}
\end{split}
\]

\noindent At the current turn, only the new user query $q_n$ is encoded into an embedding $q_n^{\text{emb}}$. This embedding is compared against the pre-computed turn-level history embeddings $\mathbf{H}^{\text{emb}}$ to obtain a relevance-score sequence $\mathbf{S}$. Using $\mathbf{S}$, our extended Kadane algorithm \texttt{KadaneDial} (Algorithm~\ref{algo:kadaneX}) identifies one or more consecutive spans in the dialogue history whose cumulative relevance remains high—for example, spans $[h_{s_1{:}e_1}], [h_{s_2{:}e_2}], [h_{s_3{:}e_3}]$, where each span corresponds to a slice of consecutive turns. These spans are then concatenated in their original chronological order to preserve dialogue continuity, forming the pruned history $\mathcal{H}' = [h_{s_1}, \dots, h_{e_1}, h_{s_2}, \dots, h_{e_2}, h_{s_3}, \dots, h_{e_3}]$. This continuity-preserving sequence $\mathcal{H}'$ is provided to the LLM agent together with the current $q_n$ as:
\[
\begin{split}
{q_n}^{\text{emb}} &= \texttt{B}(q_n) \in \mathbb{R}^{1 \times d} \\
\mathbf{S} &= \mathbf{H}^{\text{emb}} \cdot {q_n}^{\text{emb}} \in \mathbb{R}^{(n{-}1) \times 1} \\
\mathcal{H}' &= \texttt{KadaneDial}(\mathbf{S}) \\
a_n &= \texttt{LLM}([\mathcal{H}';\, q_n])
\end{split}
\]
\vspace{-0.3cm}

\SetKwComment{Comment}{$\triangleright$\ }{}
\SetKwInOut{Input}{Input}
\SetKwInOut{Output}{Output}

\begin{algorithm}[h]
\small
\caption{\textbf{.} $\;$\texttt{KadaneDial}}
\label{algo:kadaneX}
\Input{scores $\mathbf{S} = [s_1, \dots, s_{n-1}]$, \\
gain threshold $\tau$, \\
stopping threshold $\theta$}
\Output{A set of dialogue segments $\mathcal{H}'$}
\vspace{0.1cm}
$\mu \gets \frac{1}{n{-}1} \sum_{i=1}^{n{-}1} s_i$

$\sigma \gets \sqrt{\frac{1}{n{-}1} \sum_{i=1}^{n{-}1} (s_i - \mu)^2}$

$\mathbf{z} \gets [z_i]_{i=1}^{n{-}1}$ where $z_i = \frac{s_i - \mu}{\sigma}$

$\mathbf{g} \gets [g_i]_{i=1}^{n{-}1}$ where $g_i = z_i - \tau$

\vspace{0.1cm}

$\mathcal{H}' \gets \emptyset$, $\:$ $\mathbb{G} \gets +\infty$

\While{$\mathbb{G} \geq \theta$}{
    $m \gets 0$, $\;m^* \gets -\infty$, $\;i_s \gets 1$, $\;i_{\text{temp}} \gets 1$

    \For{$j = 1$ \textbf{to} $n{-}1$}{
      \If{$m + g_j > g_j$}{
        $m \gets m + g_j$
      }
      \Else{
        $m \gets g_j$, $\;i_{\text{temp}} \gets j$
      }
      \If{$m > m^*$}{
        $m^* \gets m$, $\;(i_s, i_e) \gets (i_{\text{temp}}, j)$
      }
    }

    $\mathcal{H}' \gets \mathcal{H}' \cup \{(i_s, i_e)\}$

    $g_k \gets -\infty \quad \forall k \in \{i_s, \dots, i_e\}$

    $\mathbb{G} \gets m^*$
}
\Return{$\mathcal{H}'$}
\end{algorithm}
\vspace{-0.1cm}

\noindent In \texttt{KadaneDial}, the classic Kadane’s algorithm (\texttt{L7}–\texttt{L14}) is adapted to identify salient context segments in long-form dialogue. The relevance score sequence $\mathbf{S}$ is reinterpreted as a one-dimensional signal, where consecutive high-scoring regions indicate discourse segments that are locally coherent and relevant to the current turn. To account for variation in score distributions across heterogeneous dialogues and queries, we standardize the relevance signal using z-score normalization and apply a gain shift controlled by~$\tau$ (\texttt{L1}–\texttt{L4}). Here, $z_i$ is the normalized relevance signal at turn $i$, and $g_i = z_i - \tau$ measures its excess over a threshold $\tau$: positive values increase a span’s cumulative gain, while negative values decrease it. The algorithm then iteratively identifies additional spans whose cumulative gain $\mathbb{G}$ (i.e., the score sum of the best segment in each iteration) exceeds a stopping threshold~$\theta$. This design reflects a conservative strategy that prioritizes recall, favoring the inclusion of borderline-relevant context over the risk of omitting critical context, while preserving the sequential nature of dialogue.

%% file: latex/section/experiment.tex
\subsection{Datasets}
\label{subsec:dataset}
We use three long-form open-domain conversation benchmarks: \texttt{LoCoMo}, \texttt{MT-Bench+}, and \texttt{SCM4LLMs}. Table~\ref{tab:exp_benchmark} summarizes their key statistics. \texttt{MT-Bench+} and \texttt{SCM4LLMs} consist of human–agent conversations, while \texttt{LoCoMo} features human–human dialogues. \texttt{LoCoMo} has the longest conversations, averaging 300 turns. 

\begin{table}[h!]
\small
\resizebox{\linewidth}{!}{%
\begin{tabular}{c|r|rrr}
\toprule
\textbf{Benchmark} & \multicolumn{1}{c}{\textbf{D. (\#)}} & \multicolumn{1}{|c}{\textbf{Turn (Avg.)}} & \multicolumn{1}{c}{\textbf{Token (Avg.)}} & \multicolumn{1}{c}{\textbf{Test (Avg.)}} \\ \midrule
\texttt{LoCoMo} & 10 & 301.0 & 25,750 & 198.6 \\
\texttt{MT-Bench+} & 11 & 65.5 & 20,364 & 26.2 \\
\texttt{SCM4LLMs} & 10 & 64.1 & 24,006 & 5.3 \\ \bottomrule
\end{tabular}%
}
\caption{Statistics of the three benchmarks, including the number of dialogues (D.), and the average turns, tokens, and testable QA pairs per dialogue.}
\vspace{-0.2cm}
\label{tab:exp_benchmark}
\end{table}

\noindent Since the \texttt{LoCoMo} contains a large test set with nearly 200 QA pairs per dialogue ($\approx$ 10 $\times$ 200 $=$ 2,000 in total), the cost of LLM API calls becomes a concern. To make evaluation feasible, we randomly sample 50 test QA pairs per dialogue (10 $\times$ 50 $=$ 500 in total) for our experiments~\footnote{It costs about \$20 to run and evaluate one experimental unit on 500 \texttt{LoCoMo} samples using GPT-4o.}. The test QA pairs exist separately from the dialogues, and the sampling is performed only at the level of test queries without altering the underlying dialogues.

\subsection{Models}
We evaluate \textsc{DyCP} against six methods, as follows: \emph{No Context}, \emph{Full Context}, \texttt{MemoChat}, \texttt{SCM4LLMs}, \texttt{SeCom} and \texttt{CondMem}.
\vspace{-0.2cm}
\begin{itemize}
\setlength\itemsep{-0.1cm}
 \item \emph{No Context} runs without any dialogue history, serving as a lower-bound reference. 
 \item \emph{Full Context} prepends the entire dialogue history to the input, offering a comparison point without any context management.
 \item \texttt{SCM4LLMs} \cite{wang2024enhancing} employs multiple LLM-based agents, such as memory summarization and control, in a turn-level retrieval framework. 
 \item \texttt{MemoChat} \cite{lu2023memochattuningllmsuse} manages segment-level dialogue context, using LLMs to perform both segmentation and retrieval.
 \item \texttt{SeCom} \cite{pan2025memoryconstructionretrievalpersonalized} refines dialogue context through LLM-based segmentation, dense retrieval, and context denoising to construct compact contexts.
 \item \texttt{CondMem} \cite{yuan-etal-2025-personalized} builds hybrid memory representations by combining summarization-based dialogue compression with selective memory construction.
\end{itemize}
\vspace{-0.1cm}
\noindent The main experiments were conducted using three conversational agents via API access as primary backbones: GPT-4o, Claude 3.7 (Sonnet), and GPT-4o mini. GPT-4o and Claude 3.7 are comparable models from different providers, while GPT-4o mini is a smaller variant of GPT-4o that offers moderate performance with the same input context limit. In addition, we include two recently released models (GPT-4.1 and Claude 4.0) as supplementary evaluations to assess improvements in long-context understanding and to examine whether our method remains effective. Table~\ref{tab:api_price} summarizes the per-million-token input and output costs for all API models used.

\vspace{-0.1cm}
\begin{table}[htp!]
\small
\resizebox{\linewidth}{!}{%
\begin{tabular}{l|l|r|r|r|r}
\toprule
\textbf{Model} & \textbf{provider} & \textbf{Released} & \makecell{\textbf{Context} \\ \textbf{Limit}} & \makecell{\textbf{Input} \\ \textbf{Price}} & \makecell{\textbf{Output} \\ \textbf{Price}} \\
\midrule
GPT-4o & OpenAI & 2024-05 & 128K & \$ 2.50 & \$ 10.00 \\
Claude 3.7 & Anthropic & 2025-02 & 200K & \$ 3.00 & \$ 15.00 \\ \midrule \midrule
GPT-4o mini & OpenAI & 2024-07 & 128K & \$ 0.15 & \$ $\:$ 0.60 \\ \midrule \midrule
GPT-4.1 & OpenAI & 2025-05 & 1M & \$ 2.00 & \$ $\:$ 8.00 \\
Claude 4.0 & Anthropic & 2025-05 & 200K & \$ 3.00 & \$ 15.00 \\ \bottomrule
\end{tabular}%
}
\caption{Input and output pricing (per 1M tokens) for each model and access platform, based on publicly available rates at the time of writing.}
\label{tab:api_price}
\vspace{-0.2cm}
\end{table}

\noindent We focus on closed models, where inference-level optimizations (e.g., KV cache reuse) are largely provider-managed and developers have limited control, making context management methods particularly useful for managing effective input length and cost. The limitations and broader implications of this setting are discussed in Section~\ref{sec:limitation}. 

We further examine whether \textsc{DyCP} generalizes to open-source conversational models via complementary experiments with two long-context chat models that can ingest the full dialogue histories in our benchmarks: \dotuline{Qwen2.5-7B-Instruct-1M}\footnote{\url{https://huggingface.co/Qwen/Qwen2.5-7B-Instruct-1M}} \cite{qwen2025qwen25technicalreport} and \dotuline{Mistral-Nemo-Instruct-2407}\footnote{\url{https://huggingface.co/mistralai/Mistral-Nemo-Instruct-2407}} \cite{mistral-nemo-2024}.

\vspace{0.2cm}
\subsubsection{Growth of LLM Context Windows}
\label{growth_llm_context}
\begin{figure}[h!]
    \centering
    \includegraphics[width=1.0\linewidth]{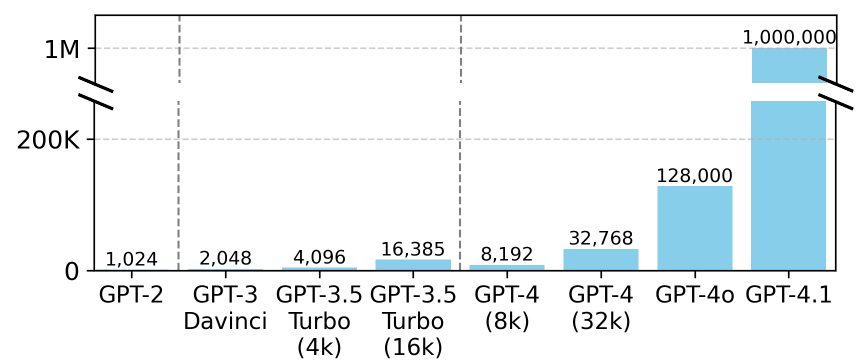}
\caption{Input context limits (in tokens) of GPT APIs, ranging from GPT-2 to GPT-4 series.}
\label{fig:gpt_context_limit}
\vspace{-0.2cm}
\end{figure}

Modern LLMs have steadily expanded their context limits over successive generations. As shown in Figure~\ref{fig:gpt_context_limit}, GPT models have grown from about 1K tokens in GPT-2 to up to 1M tokens in recent GPT-4 models. However, prior experiments such as \texttt{MemoChat}, \texttt{SCM4LLMs}, and \texttt{SeCom} were constrained to GPT-3.5–Turbo–level models (4k tokens), which cannot ingest the full dialogue history of long-form benchmarks—for example, even one-third of a single \texttt{LoCoMo} dialogue exceeds this limit. Thus, any \emph{Full Context} baseline in those settings would have required truncation. Although \texttt{CondMem} uses GPT-4 for memory construction, it was not evaluated on publicly released long-form dialogue benchmarks, and its \emph{Full Context} behavior remains unclear. In contrast, we conduct comparative experiments using modern models that support at least 128k tokens, enabling the entire dialogue history of all benchmarks to be included without truncation and allowing a direct comparison between long-context LLMs and context-management methods.

\vspace{-0.2cm}

\subsubsection{Computational Cost Analysis}
\label{para_cost}
As shown in Table~\ref{tab:llm_call}, the compared methods differ in both the number of LLM calls they require and the input scope of each call. We analyze these differences by separating \emph{offline} preprocessing cost from \emph{online} inference-time cost.

\vspace{0.2cm}
\noindent \textbf{Offline and preprocessing cost.} $\:$ Several methods require LLM calls before inference to construct or update memory. \texttt{MemoChat}, \texttt{SeCom}, and \texttt{CondMem} all rely on LLM-based memory construction that takes all previous turns (\textit{AT}) as input, making their designs appear similar at a high level. In practice, however, their preprocessing cost differs substantially in our benchmarks, where the full dialogue history is already available prior to evaluation. \texttt{MemoChat} and \texttt{SeCom} require only a single \(O(1 \times |AT|)\) pass over the dialogue. In contrast, \texttt{CondMem} constructs memory cumulatively: each update reprocesses the entire dialogue history and feeds the resulting memory into the next step. This results in \(O(n \times |AT|)\) preprocessing calls, making API-based evaluation impractical. We therefore evaluate \texttt{CondMem} only with open-source models (Section~\ref{result_opensource}). \texttt{SCM4LLMs} also performs offline preprocessing, but operates at the turn level, avoiding repeated full-history calls and remaining substantially cheaper than \texttt{CondMem}.

\vspace{0.2cm}
\noindent \textbf{Online inference cost.} $\:$ During inference, \texttt{SeCom} and \textsc{DyCP} are the most efficient methods, each requiring a single memory-and-query (\textit{M}+\textit{Q}) call per user query. Other methods introduce additional LLM calls for memory control or retrieval during inference, increasing user-facing latency. Notably, \textsc{DyCP} incurs no offline LLM calls and requires a single inference-time call per query.

\begin{table}[h!]
\centering
\small
\resizebox{1.0\linewidth}{!}{%
\begin{tabular}{c|r|r|r}
\toprule
\multicolumn{1}{c|}{}                  & \multicolumn{1}{c|}{\textbf{Preprocess}} & \multicolumn{1}{c|}{\textbf{Offline}} & \multicolumn{1}{c}{\textbf{Online}} \\ \midrule
\texttt{CondMem}      & Offline $\times$ \textit{n} &  \textit{M}$'$+\textit{Q} $\times$ 1 & \textit{M}$'$+\textit{Q} $\times$ $>$ 1 \\
                      &         & 0 | \textit{AT} $\times$ 1 & \textit{M}+\textit{Q} $\times$ 1 \\ \hline
\texttt{SCM4LLMs}     & Offline $\times$ \textit{n} & \textit{T} $\times$ 1 & \textit{Q} $\times$ 1 \\
                      &                  &            & \textit{M}+\textit{Q} $\times$ 1 \\
                      &                  &            & (\textit{AT} | \textit{M} | \textit{T})+\textit{Q} $\times$ 1 \\ \hline
\texttt{MemoChat}     & Offline $\times$ 1 & \textit{AT} $\times$ 1 & \textit{M}$'$+\textit{Q} $\times$ 1 \\
                      &                  &            & \textit{M}+\textit{Q} $\times$ 1 \\ \hline
\texttt{SeCom}        & Offline $\times$ 1 & \textit{AT} $\times$ 1 & \textit{M}+\textit{Q} $\times$ 1 \\ \hline
\textsc{DyCP}         & 0 & 0 & \textit{M}+\textit{Q} $\times$ 1 \\ 
\bottomrule
\end{tabular}%
}
\caption{
LLM call requirements for each method. 
\textbf{Preprocess} indicates offline processing applied to a dialogue before evaluating its testable queries (e.g., memory construction). \textbf{Offline} refers to per-turn LLM calls used to update memory but that do not affect user-facing latency. \textbf{Online} denotes the per-query LLM calls executed during inference. We annotate each call by its input scope: 
\textit{AT} = all previous turns, 
\textit{T} = one turn, 
\textit{M$'$} = memory candidates, 
\textit{M} = retrieved memory, 
\textit{Q} = test query. 
“$\times 1$” denotes a single call, 
“$\times >1$” denotes more than one call, 
and “$0 \mid (\cdot)$” indicates an optional call.
\textit{n} denotes the number of turns included in \textit{AT}.
}
\vspace{-0.1cm}
\label{tab:llm_call}
\end{table}


\subsubsection{Retriever}
\label{retriever}
As \textsc{DyCP}, \texttt{SCM4LLMs}, \texttt{SeCom} and \texttt{CondMem} include a retriever component in their frameworks, we experiment with four different retrievers, including variants of Contriever~\cite{DBLP:journals/corr/abs-2112-09118} and BGE~\cite{chen-etal-2024-m3}, as presented in Table~\ref{tab:result_locomo_retrieval_by_method} (Section~\ref{subsec:exp_result}). We use \dotuline{facebook/contriever-msmarco}\footnote{\url{https://huggingface.co/facebook/contriever-msmarco}} for downstream answer generation across all methods, as it consistently shows the best performance. Note that \texttt{CondMem} produces and retrieves memory chunks that do not preserve turn boundaries, making turn-level gold comparison infeasible.

\subsection{Evaluation Metrics}
\label{subsec:exp_eval}
We evaluate dialogue context management methods using three metrics: Answer Quality, Response Latency, and Retrieval Accuracy.

\vspace{0.2cm}
\noindent \textbf{Answer Quality} $\;$ We adopt GPT4Score, following prior work~\cite{lu2023memochattuningllmsuse, pan2025memoryconstructionretrievalpersonalized, yuan-etal-2025-personalized}, where GPT-4 acts as an automatic judge and assigns a score from 1 to 100 based on how well the model response answers the user question. Prior studies typically score responses using a related subset of the dialogue history as evidence for the current question. In contrast, we provide the full dialogue history up to the current turn and include the benchmark gold response as the reference for correctness (Figure~\ref{fig:prompt_gpt4score}), allowing the judge to compare against the target answer directly. All methods are evaluated under the same prompt. To assess how well GPT4Score aligns with human judgment, we conduct a small-scale human evaluation (Section~\ref{subsubsec:human_evaluation}). Given its limited scope, we treat the results as a sanity check rather than a definitive validation.

For the \texttt{LoCoMo} benchmark, which contains short, factual gold responses, we further report Exact Match (EM) and ROUGE~\cite{lin-2004-rouge}. For EM, we consider a prediction correct if the model response contains the gold response string. ROUGE is computed based on n-gram overlap between the model response and the gold response. These metrics are omitted for other benchmarks due to the length and verbosity of their gold responses, which limits the reliability of traditional lexical matching.

\begin{figure}[h!]
    \small
    \begin{tcolorbox}[width=\linewidth, colback=white, colframe=black, title=\textsc{Evaluation Prompt}]
    
    You are an impartial judge. You will be shown Conversation History, User Question, Gold Response and Model Response. \\
        
    \vspace{-0.3cm}
    
    Conversation History: \\
    \textcolor{gray}{\{\{ conversation\_history \}\}} \\
        
    \vspace{-0.3cm}
    
    User Question: \\
    \textcolor{gray}{\{\{ current\_user\_question \}\}} \\
    
    \vspace{-0.3cm}
    
    Gold Response: \\
    \textcolor{gray}{\{\{ gold\_response \}\}} \\
    
    \vspace{-0.3cm}
    
    Model Response: \\
    \textcolor{gray}{\{\{ model\_response \}\}} \\
    
    \vspace{-0.3cm}
    
    Please evaluate whether the Model Response accurately answers the User Question, referencing the proper information from the Conversation History, and using the Gold Response as a reference. Begin your evaluation by providing a short explanation, then you must rate Model Response on an integer rating of 1 to 100 by strictly following this format: [[rating]].

    \end{tcolorbox}
\vspace{-0.2cm}
\caption{Evaluation prompt used in GPT4Score}
\vspace{-0.3cm}
\label{fig:prompt_gpt4score}
\end{figure}

\vspace{0.2cm}
\noindent \textbf{Response Latency} $\;$ We measure online first-token latency—the time from receiving the current user utterance to generating the first token of the model response—to quantify the additional latency introduced by each memory management method. To observe when the first token is generated, we call all LLMs in streaming mode.

\vspace{0.2cm}
\noindent \textbf{Retrieval Accuracy} $\;$ We use Hit@k, Recall@k, and Precision@k to evaluate how accurately each method retrieves the context relevant to the current turn. As only \texttt{LoCoMo} benchmark provides ground-truth annotations of relevant context for each test question, we use \texttt{LoCoMo} for this evaluation. Each metric is defined as follows:
\vspace{-0.2cm}
\begin{itemize}
\setlength\itemsep{-0.1cm}
 \item \textbf{Hit@k} indicates whether any gold context appears in the top-k results.
 \item \textbf{Recall@k} is the proportion of gold contexts covered by the top-k results.
 \item \textbf{Precision@k} is the proportion of the top-k results that include gold context.
\end{itemize}

\subsection{Experimental Details}
\label{subsec:exp_detail}
For the implementation of \textsc{DyCP}, we set the gain threshold $\tau$ to 0.6 and the stopping threshold $\theta$ to 1.0 as hyperparameters of the \texttt{KadaneDial}. These values were chosen based on preliminary qualitative inspection of retrieved context spans and were found to produce stable pruning behavior. Because the relevance scores are z-score normalized before span selection, the thresholds operate on a standardized scale. As a result, \textsc{DyCP} is not overly sensitive to the specific numerical choices of $\tau$ and $\theta$, and we therefore use the same configuration across all datasets.

To ensure a fair comparison in retrieval-based baselines, we match the total number of retrieved turns across methods wherever applicable. For segment-level approaches such as \texttt{SeCom}, we align their retrieval volume with that of \textsc{DyCP} by matching the product of TpS and RS reported in Table~\ref{tab:result_segment}. Concretely, this corresponds to retrieving the top 5 turns on \texttt{LoCoMo}, 2 on \texttt{SCM4LLMs}, and 3 on \texttt{MT-Bench+}. \texttt{MemoChat} performs automatic segment retrieval and therefore does not require specifying a fixed $k$. For turn-level retrievers such as \texttt{SCM4LLMs} and \texttt{CondMem}, we follow the original papers and use $k{=}5$ and $k{=}3$, respectively, for all benchmarks. In all methods, retrieved items are reordered chronologically (i.e., according to their original dialogue order) before being provided to the conversational agent.

All experiments with API-based LLMs were conducted using a single NVIDIA RTX A6000 GPU to run retriever, while experiments involving open-source LLMs were run locally on four NVIDIA RTX A6000 GPUs.

\subsection{Result}
\label{subsec:exp_result}

\subsubsection{Answer Quality \& Response Latency}
\label{subsubsec:exp_quality_latency}
We evaluate \textsc{DyCP} on three long-form benchmarks in a primary API-based setting using five LLMs. As shown in Table~\ref{tab:result}, \textsc{DyCP} achieves strong answer quality while consistently exhibiting low response latency. Overall, \textsc{DyCP} matches or improves upon the \emph{Full Context} baseline in answer quality across most agents and benchmarks, with clearer gains on GPT-4o, Claude 3.7, and GPT-4o mini. As models become stronger at long-context understanding (e.g., GPT-4.1 and Claude 4.0), the quality gap between \textsc{DyCP} and \emph{Full Context} narrows, while \textsc{DyCP} continues to provide substantial efficiency gains.

As discussed in Section~\ref{growth_llm_context}, prior studies primarily evaluated LLMs with shorter context windows (e.g., 4k-token limit), where \emph{Full Context} typically required truncation. In contrast, our setting uses modern long-context LLMs that can ingest the entire history, which strengthens the \emph{Full Context} baseline and can narrow the quality gap with context-management methods. Differences in evaluation protocols (Section~\ref{subsec:exp_eval}), specifically the design of judging prompts with a stronger focus on answer correctness, may also contribute to these shifts in relative performance.

The latency benefits of \textsc{DyCP} stem from its design, which avoids additional LLM calls for memory construction or retrieval refinement, as introduced in Section~\ref{para_cost}. By minimizing both the number of LLM calls and the input size of each call, \textsc{DyCP} reduces repeated reprocessing of dialogue history and supports efficient context management for long conversations.

\begin{figure*}[h!]
    \centering
    \includegraphics[width=1.0\textwidth]{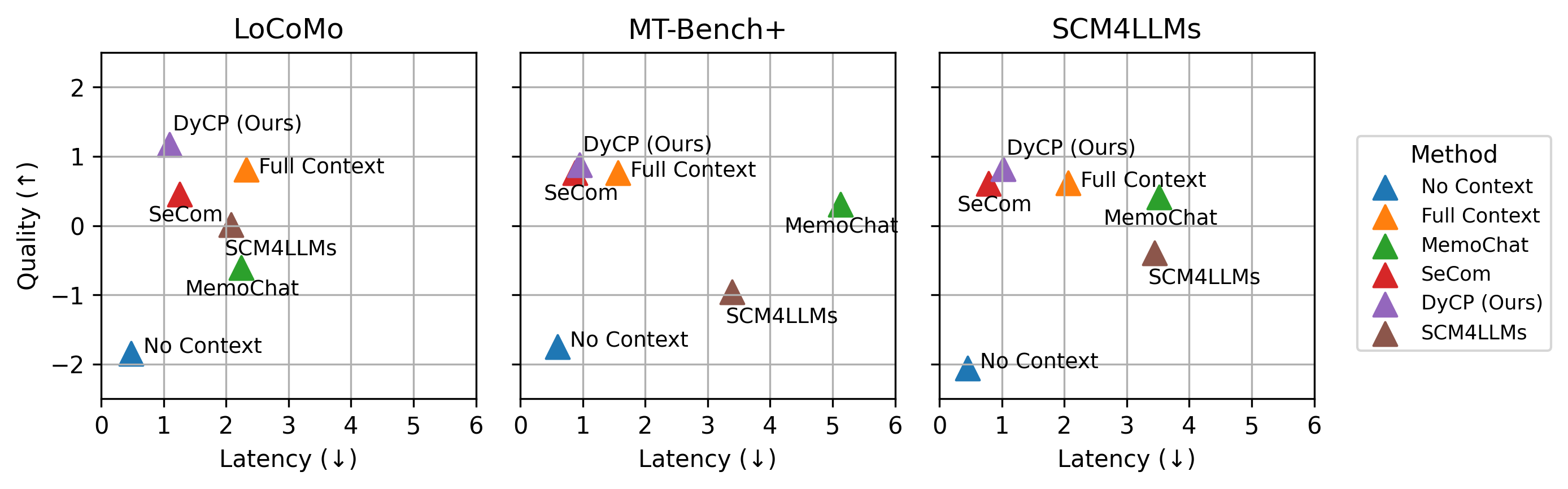}
    \vspace{-0.5cm}
\caption{
Comparison of response latency and answer quality across benchmarks using GPT-4o. The x-axis indicates first-token latency in seconds (lower is better), and the y-axis shows answer quality (higher is better). Since quality scores are on different scales across benchmarks, they are normalized as z-scores within each benchmark. \textsc{DyCP} achieves both high quality and low latency across all datasets.
}
\vspace{-0.2cm}
\label{fig:analysis_quality_latency}
\end{figure*}

Figure~\ref{fig:analysis_quality_latency} illustrates the quality--latency trade-off, where \textsc{DyCP} consistently appears in the region corresponding to relatively high quality and low latency. This trend is more noticeable on \texttt{LoCoMo}, which contains the longest conversations in terms of turns. The key distinction of our method lies in its departure from existing segmentation and retrieval paradigms. Unlike approaches based on pre-segmentation or fixed top-$k$ retrieval, \textsc{DyCP} dynamically segments and selects context for the current turn in a single pass, enabling flexible, structure-aware context selection. Further analysis is provided in Section~\ref{subsubsec:exp_retrieval}.

For the smaller GPT-4o mini model, answer quality decreases relative to GPT-4o while latency is reduced, with similar overall trends. For newer models, GPT-4.1 and Claude 4.0 show improved long-context handling; for example, on \texttt{LoCoMo}, GPT-4.1 improves over GPT-4o from 24.23 to 33.67 in EM and from 75.13 to 92.06 in GPT4Score. Under GPT-4.1, the \emph{Full Context} condition achieves performance comparable to \textsc{DyCP}. Meanwhile, \textsc{DyCP} remains more than twice as fast (0.97s vs.\ 2.41s) and uses substantially fewer input tokens (Table~\ref{tab:input_token_comparison}), indicating that selective context management remains important even as long-context LLMs improve.

\vspace{0.1cm}
\begin{table}[h!]
\centering
\small
\resizebox{0.9\linewidth}{!}{%
\begin{tabular}{c|rrr}
\toprule
             & \multicolumn{1}{c}{\textbf{LoCoMo}} & \multicolumn{1}{c}{\textbf{MT-Bench+}} & \multicolumn{1}{c}{\textbf{SCM4LLMs}} \\ \midrule
\emph{Full Context} & 25,750 & 20,364 & 24,006 \\
\textsc{DyCP} (Ours) & 4,982 & 2,698 & 4,042 \\ \bottomrule
\end{tabular}%
}
\caption{Average number of input context tokens prepended to the current user question for \emph{Full Context} and \textsc{DyCP} across benchmarks.}
\vspace{-0.3cm}
\label{tab:input_token_comparison}
\end{table}

\definecolor{customPink}{rgb}{1.0, 0.5, 0.5}
\definecolor{customGrey}{rgb}{1.0, 0.7, 0.6}
\newcommand{\linechip}[1]{\textcolor{#1}{\rule{5pt}{5pt}}}
\begin{table*}[t]
\small
\resizebox{\textwidth}{!}{%
\begin{tabular}{cc|r|rrr|r|r|r|r|r|r}
\toprule
\multicolumn{1}{l}{}                              & \multicolumn{1}{l|}{} & \multicolumn{6}{c|}{\textbf{LoCoMo}}                       & \multicolumn{2}{c|}{\textbf{MT-Bench+}}                    & \multicolumn{2}{c}{\textbf{SCM4LLMs}}                     \\ \midrule
\multicolumn{1}{c|}{\textbf{Agent}}               & \textbf{Method} & \multicolumn{1}{c|}{EM} & \multicolumn{1}{c}{R-1} & \multicolumn{1}{c}{R-2} & \multicolumn{1}{c|}{R-L} & \multicolumn{1}{c|}{Quality} & \multicolumn{1}{c|}{Latency ($\downarrow$)} & \multicolumn{1}{c|}{Quality} & \multicolumn{1}{c|}{Latency ($\downarrow$)} & \multicolumn{1}{c|}{Quality} & \multicolumn{1}{c}{Latency ($\downarrow$)} \\ \midrule
\multicolumn{1}{c|}{\multirow{6}{*}{\begin{tabular}[c]{@{}c@{}}GPT\\4o\end{tabular}}}      & \emph{No Context} & 1.79$\:\:$ & 2.73$\:\:$ & 0.33$\:\:$ & 2.36$\:\:$ & 15.57$\:\:$ {\scriptsize($\pm$ 16.1)} & 0.48 {\scriptsize($\pm$ 0.2)}  & 66.04$\:\:$ {\scriptsize($\pm$ 27.2)}     & 0.60 {\scriptsize($\pm$ 0.3)}        & 58.47$\:\:$ {\scriptsize($\pm$ 34.2)}     & 0.45 {\scriptsize($\pm$ 0.1)}       \\
\multicolumn{1}{c|}{}                             & \emph{Full Context} & \uline{24.23}$\:\:$ & 12.28$\:\:$ & \uline{5.16}$\:\:$ & 11.28$\:\:$ & \uline{75.13}$\:\:$ {\scriptsize($\pm$ 32.9)} & \cellcolor{color5} 2.32 {\scriptsize($\pm$ 0.9)} & 88.03$\:\:$ {\scriptsize($\pm$ 17.6)}     & \cellcolor{color3} 1.56 {\scriptsize($\pm$ 1.6)}        & \ul{85.51}$\:\:$ {\scriptsize($\pm$ 18.9)}     & \cellcolor{color3} 2.06 {\scriptsize($\pm$ 1.2)}       \\
\multicolumn{1}{c|}{}                             & \texttt{MemoChat} & 6.89$\:\:$ & 7.05$\:\:$ & 1.79$\:\:$ & 6.19$\:\:$ & 43.39$\:\:$ {\scriptsize($\pm$ 33.0)} & \cellcolor{color4} 2.24 {\scriptsize($\pm$ 0.8)} & 84.06$\:\:$ {\scriptsize($\pm$ 22.2)} & \cellcolor{color5} 5.12 {\scriptsize($\pm$ 2.8)} & 83.47$\:\:$ {\scriptsize($\pm$ 21.7)} & \cellcolor{color5} 3.52 {\scriptsize($\pm$ 1.8)} \\
\multicolumn{1}{c|}{}                             & \texttt{SCM4LLMs} & 12.50$\:\:$ & 9.84$\:\:$ & 3.66$\:\:$ & 9.03$\:\:$ & 49.60$\:\:$ {\scriptsize($\pm$ 37.6)} & \cellcolor{color3} 2.01 {\scriptsize($\pm$ 0.7)} & 72.98$\:\:$ {\scriptsize($\pm$ 27.0)} & \cellcolor{color4} 3.39 {\scriptsize($\pm$ 2.0)} & 76.11$\:\:$ {\scriptsize($\pm$ 27.5)} & \cellcolor{color4} 3.43 {\scriptsize($\pm$ 2.3)} \\
\multicolumn{1}{c|}{}                             & \texttt{SeCom}                 & 21.17$\:\:$ & \uline{12.45}$\:\:$ & 4.70$\:\:$ & \uline{11.32}$\:\:$ & 67.19$\:\:$ {\scriptsize($\pm$ 36.2)} & \cellcolor{color2} 1.26 {\scriptsize($\pm$ 0.7)} & \ul{88.08}$\:\:$ {\scriptsize($\pm$ 19.2)}     & \cellcolor{color1} 0.87 {\scriptsize($\pm$ 1.0)}        & 85.40$\:\:$ {\scriptsize($\pm$ 19.1)}     & \cellcolor{color1} 0.79 {\scriptsize($\pm$ 0.3)}       \\
\multicolumn{1}{c|}{} & \textsc{DyCP} (Ours) & \textbf{26.53}$\:\:$ & \textbf{14.72}\textcolor{red}{*} & \textbf{6.22}\textcolor{customGrey}{*} & \textbf{13.51}\textcolor{red}{*} & \textbf{83.27\textcolor{red}{*}} {\scriptsize($\pm$ 26.4)} & \cellcolor{color1} 1.10 {\scriptsize($\pm$ 0.5)} & \textbf{89.02}$\:\:$ {\scriptsize($\pm$ 17.1)}     & \cellcolor{color2} 0.95 {\scriptsize($\pm$ 1.0)}        & \textbf{87.57}$\:\:$ {\scriptsize($\pm$ 16.4)}     & \cellcolor{color2} 1.03 {\scriptsize($\pm$ 0.5)}       \\ \midrule
\multicolumn{1}{c|}{\multirow{6}{*}{{\begin{tabular}[c]{@{}c@{}}Claude\\3.7\end{tabular}}}}  & \emph{No Context} & 0.51$\:\:$ & 2.47$\:\:$ & 0.20$\:\:$ & 2.18$\:\:$ & 13.30$\:\:$ {\scriptsize($\pm$ $\:\;$9.5)} & 1.15 {\scriptsize($\pm$ 0.8)} & 66.87$\:\:$ {\scriptsize($\pm$ 27.8)}     & 2.12 {\scriptsize($\pm$ 1.0)}        & 57.90$\:\:$ {\scriptsize($\pm$ 36.0)}     & 1.57 {\scriptsize($\pm$ 0.7)}       \\
\multicolumn{1}{c|}{}                             & \emph{Full Context}          & \uline{26.28}$\:\:$ & \uline{6.69}$\:\:$ & \uline{2.64}$\:\:$ & \uline{5.99}$\:\:$ & \uline{74.21}$\:\:$ {\scriptsize($\pm$ 32.6)} & \cellcolor{color4} 5.54 {\scriptsize($\pm$ 1.5)} & 80.50$\:\:$ {\scriptsize($\pm$ 25.8)}     & \cellcolor{color3} 5.49 {\scriptsize($\pm$ 3.6)}        & \uline{88.08}$\:\:$ {\scriptsize($\pm$ 18.5)}     & \cellcolor{color3} 6.30 {\scriptsize($\pm$ 3.6)}       \\
\multicolumn{1}{c|}{}                             & \texttt{MemoChat} & 3.83$\:\:$ & 3.58$\:\:$ & 0.58$\:\:$ & 3.08$\:\:$ & 30.15$\:\:$ {\scriptsize($\pm$ 30.4)} & \cellcolor{color3} 5.53 {\scriptsize($\pm$ 1.3)} & 85.83$\:\:$ {\scriptsize($\pm$ 19.8)} & \cellcolor{color4} 8.18 {\scriptsize($\pm$ 2.4)} & 78.49$\:\:$ {\scriptsize($\pm$ 29.4)} & \cellcolor{color4} 7.48 {\scriptsize($\pm$ 2.3)} \\
\multicolumn{1}{c|}{} & \texttt{SCM4LLMs} & 22.19$\:\:$ & 5.80$\:\:$ & 1.86$\:\:$ & 5.25$\:\:$ & 62.82$\:\:$ {\scriptsize($\pm$ 35.1)} & \cellcolor{color5} 6.65 {\scriptsize($\pm$ 1.2)} & 76.37$\:\:$ {\scriptsize ($\pm$ 26.1)} & \cellcolor{color5} 8.63 {\scriptsize ($\pm$ 2.7)} & 79.56$\:\:$ {\scriptsize ($\pm$ 26.2)} & \cellcolor{color5} 7.82 {\scriptsize ($\pm$ 1.9)} \\
\multicolumn{1}{c|}{}                             & \texttt{SeCom}                 & 21.17$\:\:$ & 6.04$\:\:$ & 2.11$\:\:$ & 5.38$\:\:$ & 60.88$\:\:$ {\scriptsize ($\pm$ 37.4)} & \cellcolor{color2} 1.53 {\scriptsize ($\pm$ 0.6)} & \uline{87.44}$\:\:$ {\scriptsize ($\pm$ 18.7)}     & \cellcolor{color2} 3.08 {\scriptsize ($\pm$ 1.6)}        & 86.32$\:\:$ {\scriptsize ($\pm$ 18.3)}     & \cellcolor{color2} 3.54 {\scriptsize ($\pm$ 1.7)}       \\
\multicolumn{1}{c|}{}                             & \textsc{DyCP} (Ours)           & \textbf{31.89}\textcolor{customGrey}{*} & \textbf{7.98}\textcolor{red}{*} & \textbf{3.16}\textcolor{customGrey}{*} & \textbf{7.08}\textcolor{red}{*} & \textbf{83.05\textcolor{red}{*}} {\scriptsize($\pm$ 27.3)} & \cellcolor{color1} 1.52 {\scriptsize($\pm$ 0.7)} & \textbf{88.05}$\:\:$ {\scriptsize($\pm$ 18.3)}     & \cellcolor{color1} 2.93 {\scriptsize($\pm$ 1.7)}        & \textbf{88.92}$\:\:$ {\scriptsize($\pm$ 16.2)}     & \cellcolor{color1} 2.52 {\scriptsize($\pm$ 1.2)}       \\ \midrule \midrule
\multicolumn{1}{c|}{\multirow{6}{*}{{\begin{tabular}[c]{@{}c@{}}GPT\\4o\\mini\end{tabular}}}} & \emph{No Context} & 1.79$\:\:$ & 3.09$\:\:$ & 0.33$\:\:$ & 2.66$\:\:$ & 17.19$\:\:$ {\scriptsize($\pm$ 17.9)} & 0.48 {\scriptsize($\pm$ 0.5)} & 66.13$\:\:$ {\scriptsize($\pm$ 27.0)}     & 0.55 {\scriptsize($\pm$ 0.3)}        & 59.62$\:\:$ {\scriptsize($\pm$ 33.3)}     & 0.53 {\scriptsize($\pm$ 0.3)}       \\
\multicolumn{1}{c|}{}                             & \emph{Full Context}          & \uline{24.49}$\:\:$ & \uline{15.04}$\:\:$ & \uline{6.01}$\:\:$ & \uline{13.91}$\:\:$ & \uline{74.96}$\:\:$ {\scriptsize($\pm$ 30.9)} & \cellcolor{color4} 2.43 {\scriptsize($\pm$ 1.4)} & 84.88$\:\:$ {\scriptsize($\pm$ 21.6)}     & \cellcolor{color3} 1.22 {\scriptsize($\pm$ 0.9)}        & 83.08$\:\:$ {\scriptsize($\pm$ 20.5)}     & \cellcolor{color3} 1.63 {\scriptsize($\pm$ 1.4)}       \\
\multicolumn{1}{c|}{}                             & \texttt{MemoChat} & 9.69$\:\:$ & 7.69$\:\:$ & 2.65$\:\:$ & 6.83$\:\:$ & 45.62$\:\:$ {\scriptsize($\pm$ 35.1)} & \cellcolor{color5} 2.46 {\scriptsize($\pm$ 1.8)} & 74.39$\:\:$ {\scriptsize($\pm$ 26.8)} & \cellcolor{color4} 3.94 {\scriptsize($\pm$ 2.0)} & 78.42$\:\:$ {\scriptsize($\pm$ 27.6)} & \cellcolor{color4} 3.24 {\scriptsize($\pm$ 1.5)} \\
\multicolumn{1}{c|}{}                             & \texttt{SCM4LLMs} & 19.13$\:\:$ & 13.64$\:\:$ & 5.40$\:\:$ & 12.59$\:\:$ & 66.47$\:\:$ {\scriptsize($\pm$ 34.7)} & \cellcolor{color3} 2.10 {\scriptsize($\pm$ 0.6)} & 80.23$\:\:$ {\scriptsize($\pm$ 23.8)} & \cellcolor{color5} 4.97 {\scriptsize($\pm$ 3.1)} & \uline{84.38}$\:\:$ {\scriptsize($\pm$ 18.5)} & \cellcolor{color5} 4.22 {\scriptsize($\pm$ 3.0)} \\
\multicolumn{1}{c|}{}                             & \texttt{SeCom}                 & 19.64$\:\:$ &  12.01$\:\:$ & 4.50$\:\:$ &  10.97$\:\:$ & 66.09$\:\:$ {\scriptsize($\pm$ 34.9)} & \cellcolor{color2} 0.88 {\scriptsize($\pm$ 0.5)} & \uline{86.52}$\:\:$ {\scriptsize($\pm$ 20.0)}     & \cellcolor{color2} 0.77 {\scriptsize($\pm$ 1.0)}        & 83.09$\:\:$ {\scriptsize($\pm$ 22.1)}     & \cellcolor{color1} 0.74 {\scriptsize($\pm$ 0.2)}       \\
\multicolumn{1}{c|}{}                             & \textsc{DyCP} (Ours)           & \textbf{28.32}$\:\:$ & \textbf{16.12}$\:\:$ & \textbf{6.69}$\:\:$ & \textbf{14.79}$\:\:$ & \textbf{80.44\textcolor{customGrey}{*}} {\scriptsize($\pm$ 28.5)} & \cellcolor{color1} 0.88 {\scriptsize($\pm$ 0.4)} & \textbf{87.24}$\:\:$ {\scriptsize($\pm$ 19.3)}     & \cellcolor{color1} 0.75 {\scriptsize($\pm$ 0.4)}        & \textbf{84.55}$\:\:$ {\scriptsize($\pm$ 18.1)}     & \cellcolor{color2} 0.99 {\scriptsize($\pm$ 0.7)}       \\ \midrule \midrule
\multicolumn{1}{c|}{\multirow{3}{*}{{\begin{tabular}[c]{@{}c@{}}GPT\\4.1\end{tabular}}}} & \emph{No Context} & 1.79$\:\:$ & 4.76$\:\:$ & 0.66$\:\:$ & 3.95$\:\:$ & 25.21$\:\:$ {\scriptsize($\pm$ 24.9)} & 0.43 {\scriptsize($\pm$ 0.3)} & 67.54$\:\:$ {\scriptsize($\pm$ 28.0)} & 0.46 {\scriptsize($\pm$ 0.3)} & 62.41$\:\:$ {\scriptsize($\pm$ 36.5)} & 0.43 {\scriptsize($\pm$ 0.1)} \\
\multicolumn{1}{c|}{} & \emph{Full Context} & \uline{33.67}$\:\:$ & \uline{12.31}$\:\:$ & \uline{5.63}$\:\:$ & \uline{11.35}$\:\:$ & \textbf{92.06}$\:\:$ {\scriptsize($\pm$ 17.4)} & \cellcolor{color3} 2.41 {\scriptsize($\pm$ 1.7)} & \textbf{90.50}$\:\:$ {\scriptsize($\pm$ 15.7)} & \cellcolor{color3} 1.63 {\scriptsize($\pm$ 1.6)} & \uline{90.83}$\:\:$ {\scriptsize($\pm$ 14.1)} & \cellcolor{color3} 1.91 {\scriptsize($\pm$ 1.3)} \\
\multicolumn{1}{c|}{} & \textsc{DyCP} (Ours) & \textbf{35.46}$\:\:$ & \textbf{12.45}$\:\:$ & \textbf{5.91}$\:\:$ & \textbf{11.56}$\:\:$ & \uline{91.46}$\:\:$ {\scriptsize($\pm$ 18.3)} & \cellcolor{color1} 0.97 {\scriptsize($\pm$ 0.5)} & \uline{90.18}$\:\:$ {\scriptsize($\pm$ 16.4)} & \cellcolor{color1} 0.71 {\scriptsize($\pm$ 0.7)} & \textbf{91.08}$\:\:$ {\scriptsize($\pm$ 14.6)} & \cellcolor{color1} 0.77 {\scriptsize($\pm$ 0.3)} \\ \midrule
\multicolumn{1}{c|}{\multirow{3}{*}{{\begin{tabular}[c]{@{}c@{}}Claude\\4.0\end{tabular}}}} & \emph{No Context} & 0.77$\:\:$ & 2.83$\:\:$ & 0.21$\:\:$ & 2.46$\:\:$ & 13.30$\:\:$ {\scriptsize($\pm$ $\:\;$9.5)} & 1.15 {\scriptsize($\pm$ 0.8)} & 66.87$\:\:$ {\scriptsize($\pm$ 27.8)} & 2.12 {\scriptsize($\pm$ 1.0)} & 57.89$\:\:$ {\scriptsize($\pm$ 36.0)} & 1.57 {\scriptsize($\pm$ 0.7)} \\
\multicolumn{1}{c|}{} & \emph{Full Context} & \uline{34.69}$\:\:$ & \uline{6.89}$\:\:$ & \uline{2.98}$\:\:$ & \uline{6.24}$\:\:$ & \uline{89.57}$\:\:$ {\scriptsize($\pm$ 19.9)} & \cellcolor{color3} 5.30 {\scriptsize($\pm$ 1.7)} & \uline{85.71}$\:\:$ {\scriptsize($\pm$ 20.4)} & \cellcolor{color3} 2.97 {\scriptsize($\pm$ 0.7)} & \uline{84.89}$\:\:$ {\scriptsize($\pm$ 22.0)} & \cellcolor{color3} 3.25 {\scriptsize($\pm$ 0.7)} \\
\multicolumn{1}{c|}{} & \textsc{DyCP} (Ours) & \textbf{37.24}\textcolor{customGrey}{*} & \textbf{7.48}$\:\:$ & \textbf{3.41}$\:\:$ & \textbf{6.83}$\:\:$ & \textbf{90.45}$\:\:$ {\scriptsize($\pm$ 19.2)} & \cellcolor{color1} 1.85 {\scriptsize($\pm$ 0.5)} & \textbf{88.39}\textcolor{customGrey}{*} {\scriptsize($\pm$ 17.5)} & \cellcolor{color1} 1.55 {\scriptsize($\pm$ 0.4)} & \textbf{88.19}$\:\:$ {\scriptsize($\pm$ 16.0)} & \cellcolor{color1} 1.79 {\scriptsize($\pm$ 0.5)} \\ \bottomrule
\end{tabular}%
}
\caption{
Evaluation of answer quality ($\uparrow$) and response latency ($\downarrow$) on three long-form dialogue benchmarks: \texttt{LoCoMo}, \texttt{MT-Bench+}, and \texttt{SCM4LLMs}, using five conversational agents (GPT-4o, Claude 3.7, GPT-4o mini, GPT-4.1, and Claude 4.0). We primarily use GPT4Score as the quality metric. For \texttt{LoCoMo}, which features short, factual gold answers, we also report lexical-based quality metrics such as Exact Match (EM) and ROUGE-1/2/L, as they provide a reliable signal of answer quality in this setting. The best score per experimental unit is \textbf{bolded}, and the second-best is \underline{underlined}. We also perform paired t-tests between these top two methods to assess whether the best scores are statistically significantly higher than the second-best scores: \textcolor{red}{*} for $p < 0.01$, \textcolor{customPink}{*} for $p < 0.05$ and \textcolor{customGrey}{*} for $p < 0.1$. Latency is reported in seconds, with darker cells indicating higher latency: \linechip{color5} High $\;$ \linechip{color3} Medium $\;$ \linechip{color1} Low (preferred).
}
\vspace{-0.1cm}
\label{tab:result}
\end{table*}

\begin{table*}[h!]
\small
\resizebox{\textwidth}{!}{%
\begin{tabular}{c|c|rrrr|rrrr|rrrr}
\toprule
\textbf{Method} & \textbf{Retriever} & \multicolumn{1}{c}{\textbf{H@1}} & \multicolumn{1}{c}{\textbf{H@3}} & \multicolumn{1}{c}{\textbf{H@5}} & \multicolumn{1}{c|}{\textbf{H}} & \multicolumn{1}{c}{\textbf{R@1}} & \multicolumn{1}{c}{\textbf{R@3}} & \multicolumn{1}{c}{\textbf{R@5}} & \multicolumn{1}{c|}{\textbf{R}} & \multicolumn{1}{c}{\textbf{P@1}} & \multicolumn{1}{c}{\textbf{P@3}} & \multicolumn{1}{c}{\textbf{P@5}} & \multicolumn{1}{c}{\textbf{P}} \\ 
\midrule
\multirow{4}{*}{\texttt{SCM4LLMs}} 
  & facebook/contriever & 0.3529 & 0.5473 & 0.6650 & 0.7647 & 0.3005 & 0.4779 & 0.5831 & 0.6839 & 0.3529 & 0.1969 & 0.1473 & 0.0908 \\
  & facebook/contriever-msmarco & \textbf{0.4194} & \textbf{0.6445} & \textbf{0.7187} & \textbf{0.8005} & \textbf{0.3663} & \textbf{0.5708} & \textbf{0.6380} & \textbf{0.7176} & \textbf{0.4194} & \textbf{0.2344} & \textbf{0.1621} & \textbf{0.0939} \\
  & BAAI/bge-base-en & 0.3708 & 0.6087 & 0.6829 & 0.7494 & 0.3168 & 0.5319 & 0.5983 & 0.6693 & 0.3708 & 0.2234 & 0.1540 & 0.0900 \\
  & BAAI/bge-base-en-v1.5 & 0.3836 & 0.6292 & 0.6880 & 0.7801 & 0.3311 & 0.5502 & 0.6056 & 0.6952 & 0.3836 & 0.2285 & 0.1575 & 0.0928 \\
\midrule
\multirow{4}{*}{\texttt{SeCom}} 
  & facebook/contriever & 0.2916 & 0.5192 & 0.6650 & 0.8414 & 0.2439 & 0.4344 & 0.5805 & 0.7825 & 0.2916 & 0.1876 & 0.1524 & 0.1066 \\
  & facebook/contriever-msmarco & \textbf{0.3325} & \textbf{0.5601} & \textbf{0.6726} & \textbf{0.8542} & \textbf{0.2788} & \textbf{0.4822} & \textbf{0.5886} & \textbf{0.7977} & \textbf{0.3325} & \textbf{0.2046} & \textbf{0.1540} & \textbf{0.1097} \\
  & BAAI/bge-base-en  & 0.2737 & 0.5320 & 0.6471 & 0.8389 & 0.2090 & 0.4520 & 0.5708 & 0.7844 & 0.2737 & 0.1969 & 0.1499 & 0.1079 \\
  & BAAI/bge-base-en-v1.5 & 0.3171 & 0.5550 & 0.6675 & 0.8261 & 0.2548 & 0.4737 & 0.5846 & 0.7732 & 0.3171 & 0.2038 & 0.1535 & 0.1066 \\
\midrule
\multirow{4}{*}{\textsc{DyCP} (Ours)} 
  & facebook/contriever & 0.5115 & 0.7903 & 0.8670 & 0.9488 & 0.4613 & 0.7143 & 0.7995 & 0.8985 & 0.5115 & 0.2847 & 0.1969 & \textbf{0.1031} \\
  & facebook/contriever-msmarco & \textbf{0.5320} & \textbf{0.7954} & \textbf{0.8849} & \textbf{0.9565} & \textbf{0.4803} & \textbf{0.7263} & \textbf{0.8179} & \textbf{0.9088} & \textbf{0.5320} & \textbf{0.2907} & \textbf{0.2005} & 0.0992 \\
  & BAAI/bge-base-en & 0.4885 & 0.7263 & 0.8465 & 0.9309 & 0.4248 & 0.6591 & 0.7710 & 0.8858 & 0.4885 & 0.2634 & 0.1893 & 0.0929 \\
  & BAAI/bge-base-en-v1.5 & 0.5243 & 0.7724 & 0.8619 & 0.9540 & 0.4594 & 0.6969 & 0.7912 & 0.9079 & 0.5243 & 0.2788 & 0.1939 & 0.0966 \\
\bottomrule
\end{tabular}%
}
\caption{
Retrieval performance on the \texttt{LoCoMo} benchmark using three retriever-based methods (\texttt{SCM4LLMs}, \texttt{SeCom}, and \textsc{DyCP} with four retrievers. We report Hit@k (H@k), Recall@k (R@k), and Precision@k (P@k) for $\text{k} = 1, 3, 5$, along with metrics computed over the full set of retrieved results by each method (denoted as H, R, P). For \textsc{DyCP}, the number of retrieved results is determined by the method (refer to Table~\ref{tab:result_segment}), whereas for \texttt{SCM4LLMs} and \texttt{SeCom}, the top-k is fixed to 10.
}
\vspace{-0.2cm}
\label{tab:result_locomo_retrieval_by_method}
\end{table*}

\subsubsection{Human Evaluation}
\label{subsubsec:human_evaluation}
\begin{figure}[h!]
    \centering
    \includegraphics[width=1.0\linewidth]{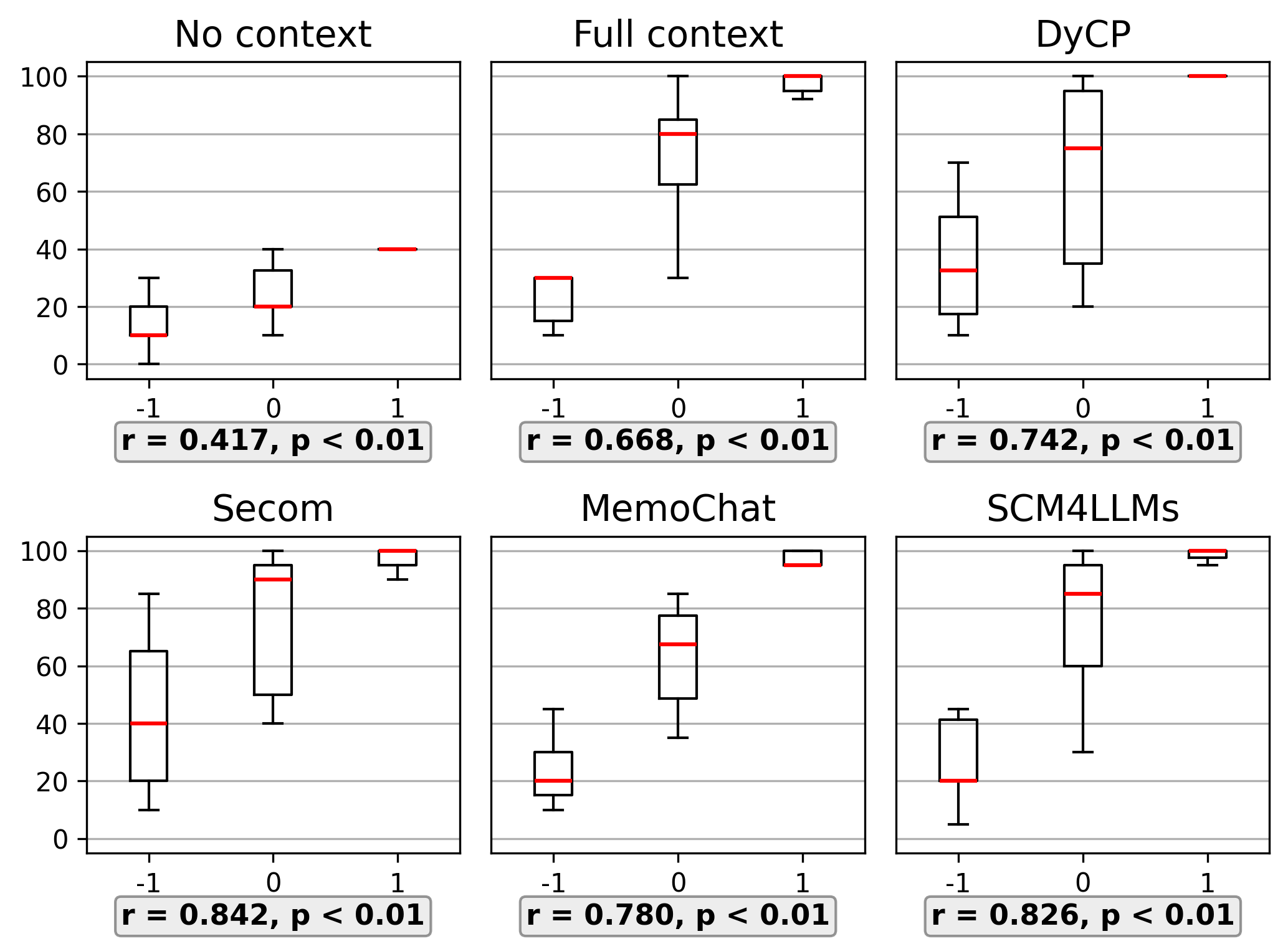}
\caption{
Comparison of GPT4Score with human correctness labels on one \texttt{LoCoMo} dialogue (50 test QA pairs) across six methods. GPT4Score (0–100) is grouped by human ratings—incorrect ($-1$), partially correct ($0$), and correct ($1$).
}
\vspace{-0.1cm}
\label{fig:human_alignment}
\end{figure}

To examine whether GPT4Score broadly aligns with human judgments, we conducted a small-scale human evaluation on a single \texttt{LoCoMo} dialogue with 50 test QA pairs. One author annotated responses from six methods: \textit{No Context}, \textit{Full Context}, \textsc{DyCP}, \texttt{SeCom}, \texttt{MemoChat}, and \texttt{SCM4LLMs}. For each question, the annotator was shown the user query, gold response, and model-generated response, with method identities hidden, and assigned a correctness label from $\{-1, 0, 1\}$, where $-1$ denotes incorrect, $0$ partially correct\footnote{For example, when the query is \textit{"When did Caroline join a mentorship program?"} and the gold answer is \textit{"The weekend before 17 July 2023"}, the response \textit{"around July 17, 2023"} captures the approximate time period but misses the exact reference.}, and $1$ correct. As shown in Figure~\ref{fig:human_alignment}, GPT4Score exhibits consistent shifts in score distributions across the three labels and shows moderate to strong correlations with human judgments across methods ($r=0.417$–$0.842$, $p < 0.01$). Given the limited scale and single-annotator setup, we interpret these results as indicative rather than definitive, suggesting that GPT4Score serves as a reasonable proxy for human evaluation in our experimental setting.

\subsubsection{Retrieval Accuracy}
\label{subsubsec:exp_retrieval}
To better understand how each context management method identifies relevant context, we compare their retrieval accuracy in Table~\ref{tab:result_locomo_retrieval_by_method} among methods that incorporate a retriever. \textsc{DyCP} consistently outperforms both \texttt{SCM4LLMs} and \texttt{SeCom}, regardless of the retriever used. Interestingly, while \texttt{SCM4LLMs} achieves slightly higher retrieval accuracy than \texttt{SeCom} (both using the top 5 retrieved results on the \texttt{LoCoMo} benchmark), \texttt{SeCom} demonstrates comparable or even better answer quality, as shown in Table~\ref{tab:result}. This suggests that segment-level retrieval—which better preserves the sequential nature and coherence of the previous context—may contribute to improved downstream generation quality.

A key difference among segment-level methods lies in how segment boundaries are constructed. \texttt{SeCom} and \texttt{MemoChat} segment the dialogue \emph{offline} before the current turn, which limits query adaptivity and can merge relevant and irrelevant turns within the same segment. In contrast, \textsc{DyCP} constructs segments \emph{on demand} based on relevance dynamics, allowing boundaries to shift with the current query. Table~\ref{tab:result_segment} reports descriptive statistics of segmentation and retrieval behavior (Turns per Segment, TpS; Retrieved Segments, RS). \textsc{DyCP} tends to use finer-grained segmentation on long, multi-topic dialogues (e.g., \texttt{LoCoMo}) and adjusts retrieval volume more flexibly across turns. TpS is not uniformly smaller across datasets: when relevant evidence spans a longer coherent stretch (e.g., in \texttt{SCM4LLMs}), \textsc{DyCP} may retain longer contiguous spans rather than forcing shorter segments. Retrieval accuracy (Table~\ref{tab:result_locomo_retrieval_by_method}) directly measures segment–query alignment, while answer quality (Table~\ref{tab:result}) captures its downstream effect.

\vspace{-0.1cm}
\begin{table}[htp!]
\centering
\small
\resizebox{\linewidth}{!}{%
\begin{tabular}{c|rr|rr|rr}
\toprule
            & \multicolumn{2}{c|}{\textbf{LoCoMo}}                & \multicolumn{2}{c|}{\textbf{MT-Bench+}}             & \multicolumn{2}{c}{\textbf{SCM4LLMs}}              \\ \midrule
            & \multicolumn{1}{c}{TpS.} & \multicolumn{1}{c|}{RS.} & \multicolumn{1}{c}{TpS.} & \multicolumn{1}{c|}{RS.} & \multicolumn{1}{c}{TpS.} & \multicolumn{1}{c}{RS.} \\ \midrule
\texttt{MemoChat}    & 21.6 & 1.3  & 3.1                      & 1.2                      & 3.2                      & 1.7                     \\
\texttt{SeCom}       & 13.2 & 5                        & 3.0                      & 3                        & 4.5                      & 2                       \\
\textsc{DyCP} (Ours) & 4.4 & 13.0 & 3.8                      & 2.2                      & 4.9                      & 2.3                     \\ \bottomrule
\end{tabular}%
}
\caption{Comparison of segment-level approaches in terms of average Turns per Segment (TpS.) and Retrieved Segments (RS.) across three benchmarks. TpS. reflects segmentation granularity, while RS. indicates the number of segments retrieved per turn. RS. for \texttt{SeCom} is manually set per benchmark, unlike the automatic retrieval used in \texttt{MemoChat} and \textsc{DyCP}. Segmentation in \texttt{MemoChat} and \texttt{SeCom} is performed by GPT-4o.}
\vspace{-0.4cm}
\label{tab:result_segment}
\end{table}

\begin{table*}[h!]
\small
\resizebox{\textwidth}{!}{%
\begin{tabular}{cc|r|rrr|r|r|r|r|r|r}
\toprule
\multicolumn{1}{l}{}                              & \multicolumn{1}{l|}{} & \multicolumn{6}{c|}{\textbf{LoCoMo}}                       & \multicolumn{2}{c|}{\textbf{MT-Bench+}}                    & \multicolumn{2}{c}{\textbf{SCM4LLMs}}                     \\ \midrule
\multicolumn{1}{c|}{\textbf{Agent}}               & \textbf{Method} & \multicolumn{1}{c|}{EM} & \multicolumn{1}{c}{R-1} & \multicolumn{1}{c}{R-2} & \multicolumn{1}{c|}{R-L} & \multicolumn{1}{c|}{Quality} & \multicolumn{1}{c|}{Latency ($\downarrow$)} & \multicolumn{1}{c|}{Quality} & \multicolumn{1}{c|}{Latency ($\downarrow$)} & \multicolumn{1}{c|}{Quality} & \multicolumn{1}{c}{Latency ($\downarrow$)} \\ \midrule
\multicolumn{1}{c|}{\multirow{4}{*}{{\begin{tabular}[c]{@{}c@{}}Qwen2.5-7B-Instruct\\(1M)\end{tabular}}}} & \emph{No Context} & 1.28$\:\:$ & 3.23$\:\:$ & 0.31$\:\:$ & 2.80$\:\:$ & 21.49$\:\:$ {\scriptsize($\pm$ 21.0)} & 0.04 {\scriptsize($\pm$ 0.0)} & 62.81$\:\:$ {\scriptsize($\pm$ 27.1)} & 0.04 {\scriptsize($\pm$ 0.0)} & 58.57$\:\:$ {\scriptsize($\pm$ 32.7)} & 0.05 {\scriptsize($\pm$ 0.1)} \\
\multicolumn{1}{c|}{} & \emph{Full Context} & \uline{22.96}$\:\:$ & 9.02$\:\:$ & 3.90$\:\:$ & 8.30$\:\:$ & \uline{72.57}$\:\:$ {\scriptsize($\pm$ 31.4)} & \cellcolor{color3} 6.56 {\scriptsize($\pm$ 1.5)} & \uline{81.47}$\:\:$ {\scriptsize($\pm$ 24.7)} & \cellcolor{color3} 3.47 {\scriptsize($\pm$ 0.6)} & \uline{74.92}$\:\:$ {\scriptsize($\pm$ 26.3)} & \cellcolor{color3} 4.41 {\scriptsize($\pm$ 1.1)} \\
\multicolumn{1}{c|}{} & \texttt{CondMem} & 16.84$\:\:$ & \textbf{12.39}\textcolor{red}{*} & \textbf{4.83}$\:\:$ & \textbf{11.49}\textcolor{red}{*} & 57.07$\:\:$ {\scriptsize($\pm$ 36.9)} & \cellcolor{color5} 7.45 {\scriptsize($\pm$ 4.7)} & 60.89$\:\:$ {\scriptsize($\pm$ 31.8)} & \cellcolor{color5} 11.36 {\scriptsize($\pm$ 7.6)} & 60.08$\:\:$ {\scriptsize($\pm$ 35.4)} & \cellcolor{color5} 10.36 {\scriptsize($\pm$ 6.8)} \\
\multicolumn{1}{c|}{} & \textsc{DyCP} (Ours) & \textbf{25.00}\textcolor{red}{*} & \uline{9.63}$\:\:$ & \uline{4.15}$\:\:$ & \uline{8.88}$\:\:$ & \textbf{75.96}$\:\:$ {\scriptsize($\pm$ 30.0)} & \cellcolor{color1} 0.98 {\scriptsize($\pm$ 0.3)} & \textbf{84.09}$\:\:$ {\scriptsize($\pm$ 22.0)} & \cellcolor{color1} 0.40 {\scriptsize($\pm$ 0.3)} & \textbf{75.62}$\:\:$ {\scriptsize($\pm$ 26.0)} & \cellcolor{color1} 0.77 {\scriptsize($\pm$ 0.5)} \\ \midrule
\multicolumn{1}{c|}{\multirow{4}{*}{{\begin{tabular}[c]{@{}c@{}}Mistral-Nemo-Instruct\\(128k)\end{tabular}}}} & \emph{No Context} & 4.08$\:\:$ & 5.24$\:\:$ & 0.92$\:\:$ & 4.64$\:\:$ & 23.90$\:\:$ {\scriptsize($\pm$ 26.4)} & 0.05 {\scriptsize($\pm$ 0.0)} & 60.82$\:\:$ {\scriptsize($\pm$ 28.7)} & 0.06 {\scriptsize($\pm$ 0.0)} & 55.28$\:\:$ {\scriptsize($\pm$ 32.4)} & 0.07 {\scriptsize($\pm$ 0.1)} \\
\multicolumn{1}{c|}{} & \emph{Full Context} & \uline{21.17}$\:\:$ & \uline{20.59}$\:\:$ & \uline{8.81}$\:\:$ & \uline{19.33}$\:\:$ & \uline{66.75}$\:\:$ {\scriptsize($\pm$ 35.4)} & \cellcolor{color5} 10.18 {\scriptsize($\pm$ 2.2)} & \uline{74.50}$\:\:$ {\scriptsize($\pm$ 26.5)} & \cellcolor{color3} 5.77 {\scriptsize($\pm$ 1.1)} & \uline{71.06}$\:\:$ {\scriptsize($\pm$ 27.7)} & \cellcolor{color5} 7.30 {\scriptsize($\pm$ 1.8)} \\
\multicolumn{1}{c|}{} & \texttt{CondMem} & 16.58$\:\:$ & 16.53$\:\:$ & 5.96$\:\:$ & 15.19$\:\:$ & 56.36$\:\:$ {\scriptsize($\pm$ 37.7)} & \cellcolor{color3} 3.25 {\scriptsize($\pm$ 1.2)} & 58.71$\:\:$ {\scriptsize($\pm$ 29.4)} & \cellcolor{color5} 7.72 {\scriptsize($\pm$ 4.7)} & 65.10$\:\:$ {\scriptsize($\pm$ 31.0)} & \cellcolor{color3} 5.57 {\scriptsize($\pm$ 3.2)} \\
\multicolumn{1}{c|}{} & \textsc{DyCP} (Ours) & \textbf{25.26}$\:\:$ & \textbf{23.44}\textcolor{customPink}{*} & \textbf{10.39}$\:\:$ & \textbf{21.84}\textcolor{customGrey}{*} & \textbf{79.04}\textcolor{red}{*} {\scriptsize($\pm$ 31.1)} & \cellcolor{color1} 1.52 {\scriptsize($\pm$ 0.5)} & \textbf{84.01}\textcolor{red}{*} {\scriptsize($\pm$ 22.4)} & \cellcolor{color1} 0.66 {\scriptsize($\pm$ 0.5)} & \textbf{80.68}\textcolor{customPink}{*} {\scriptsize($\pm$ 22.4)} & \cellcolor{color1} 1.15 {\scriptsize($\pm$ 0.6)} \\ \bottomrule
\end{tabular}%
}
\caption{
Evaluation of answer quality ($\uparrow$) and response latency ($\downarrow$) on three long-form dialogue benchmarks: \texttt{LoCoMo}, \texttt{MT-Bench+}, and \texttt{SCM4LLMs}, using two open-source long-context models (Qwen2.5-7B-Instruct, Mistral-Nemo-Instruct). All other formatting conventions follow Table~\ref{tab:result} (e.g., marking best and second-best results, $p$-value annotations, and latency shading).
}
\vspace{-0.1cm}
\label{tab:result_opensource}
\end{table*}

\subsection{Open-Source LLMs Evaluation}
\label{result_opensource}
We further examine whether \textsc{DyCP} generalizes to open-source long-context models. We also include \texttt{CondMem} in this setting; as discussed in Section~\ref{para_cost}, its preprocessing requires extensive LLM calls, making large-scale API-based evaluation costly in practice. We conduct experiments on Qwen2.5-7B-Instruct (1M context) and Mistral-Nemo-Instruct (128k context). 

As shown in Table~\ref{tab:result_opensource}, \textsc{DyCP} consistently achieves the strong answer quality across all three benchmarks, outperforming both \emph{Full Context} and \texttt{CondMem}. While \textsc{DyCP} maintains low and stable online latency, \texttt{CondMem} shows substantially higher latency with large variance due to its iterative retrieval-refinement process. Although \texttt{CondMem} reports high ROUGE-1/2/L scores on Qwen, these gains are largely driven by keyword overlap: the method often reproduces short 1–2-gram fragments from the gold answer without providing the correct factual content. This inflates ROUGE despite low semantic accuracy, as reflected in both EM and GPT4Score. These results demonstrate that \textsc{DyCP} remains effective across models, including smaller open-source LLMs. 

While locally deployed open-source models may benefit from reusing cached KV states across previous turns, many practical serving setups still recompute full-history prompts (e.g., when the model is accessed via an internal API). In such cases, \textsc{DyCP} can be particularly useful by reducing the effective input length. We further discuss these caching considerations in Section~\ref{sec:limitation}. Here, we show that \textsc{DyCP} generalizes to open-source long-context models.

%% file: latex/section/analysis.tex
To gain insight into when and how context pruning improves long-form dialogue, we analyze three factors. First, we compare GPT-4o and GPT-4.1 to assess how enhanced long-context capabilities in recent LLMs affect the benefits of pruning. We attribute observed shifts in performance to dialogue-level characteristics, such as dialogue length and the position of relevant context (Section~\ref{subsec:dialogue_characteristics}). Second, to understand what makes pruning effective, we examine different retrieval configurations, focusing on the trade-off between recall and precision (Section~\ref{subsec:recall_precision}). Third, we investigate whether continuity-preserving turns—those with low but positive relevance—meaningfully contribute to answer quality. To this end, we perform an ablation study that progressively removes the least-relevant turns within the segments selected by \textsc{DyCP}, analyzing how systematically dropping these turns affects downstream performance (Section~\ref{subsec:ablation}). All analyses in this section are conducted on the \texttt{LoCoMo} benchmark.

\subsection{A Study Across GPT-4o and GPT-4.1}
\label{subsec:dialogue_characteristics}
\begin{figure}[h!]
    \centering
    \includegraphics[width=1.0\linewidth]{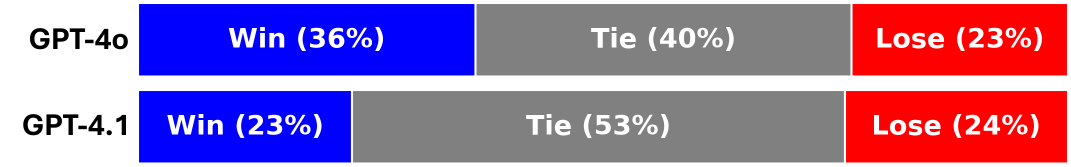}
\caption{
Win–Tie–Lose breakdown of \textsc{DyCP} compared to \emph{Full Context}. Each segment represents the proportion of test cases where our method achieved a higher score (\textcolor{blue}{Win}), the same score (\textcolor{darkgray}{Tie}), or a lower score (\textcolor{red}{Lose}).
}
\label{fig:win_lose}
\vspace{-0.1cm}
\end{figure}

Figure~\ref{fig:win_lose} shows that \textsc{DyCP} achieves better answer quality than \emph{Full Context} in a larger portion of cases (\textcolor{blue}{Win}) under GPT-4o. However, with GPT-4.1, the two methods produce more similar results, as the proportion of ties increases and the wins decrease. \textsc{DyCP} still offers substantial gains in efficiency, such as fewer input tokens and lower latency (Section~\ref{subsubsec:exp_quality_latency}), but it remains important to understand what accounts for the improved long-context processing observed in GPT-4.1.

\begin{figure}[h!]
    \centering
    \includegraphics[width=1.0\linewidth]{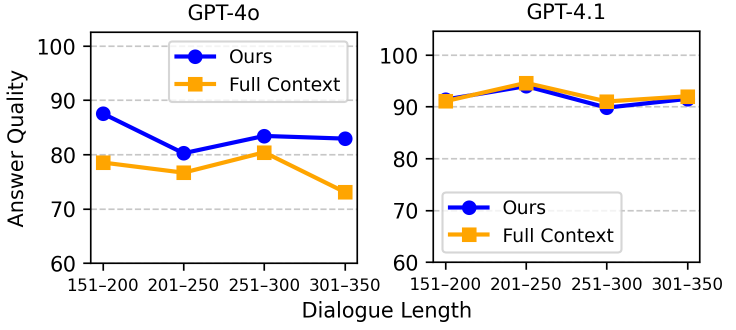}
\caption{
Comparison of answer quality between \textsc{DyCP} and \emph{Full Context} based on the number of preceding dialogue turns for GPT-4o and GPT-4.1.
}
\label{fig:analysis_length}
\vspace{-0.1cm}
\end{figure}

\noindent Figure~\ref{fig:analysis_length} shows how the number of preceding dialogue turns affects the answer quality in the current turn. Under GPT-4o, \emph{Full Context} performance drops sharply for long conversations—particularly those exceeding 300 turns—whereas \textsc{DyCP} maintains stable performance even in long conversations. In contrast, GPT-4.1 exhibits no such degradation, with \emph{Full Context} performing comparably to \textsc{DyCP} across all dialogue lengths.

\begin{figure}[h!]
    \centering
    \includegraphics[width=1.0\linewidth]{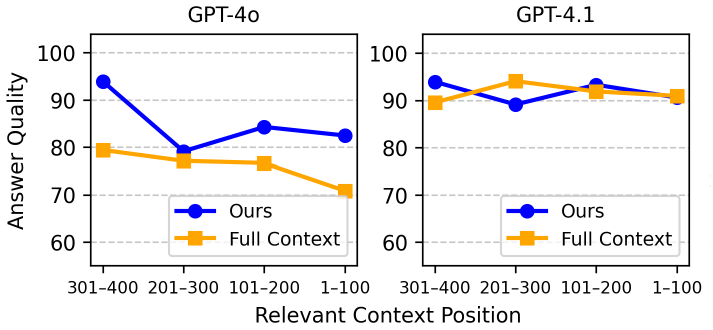}
\caption{
Comparison of answer quality between \textsc{DyCP} and \emph{Full Context} based on the relevant context positions for GPT-4o and GPT-4.1.
}
\label{fig:analysis_position}
\vspace{-0.1cm}
\end{figure}

\noindent Figure~\ref{fig:analysis_position} shows the impact of relevant context position on answer quality. Under GPT-4o, \emph{Full Context} exhibits strong position sensitivity, performing worse when the relevant information appears early in the input (1--100), and gradually improving as it moves closer to the end (301--400). This indicates a recency bias \citep{peysakhovich2023attentionsortingcombatsrecency, press2022trainshorttestlong}, where earlier content is underutilized. With GPT-4.1, position sensitivity largely disappears for \emph{Full Context}, and its performance becomes comparable to that of \textsc{DyCP}, suggesting improved capacity to retain and attend to long-range information.

\noindent When \emph{Full Context} setting fails under GPT-4o, it typically shows one of two behaviors: (1) hallucinating information or (2) incorrectly claiming that no relevant information is available. Table~\ref{fig:full_failure_1} and Table~\ref{fig:full_failure_2} (\ref{subsec:example}) illustrate examples of each case. On the other hand, when \emph{Full Context} yields better results than \textsc{DyCP} (\textcolor{red}{Lose}), this is primarily attributable to limitations in retrieval performance, which directly affect \textsc{DyCP}. A more detailed error analysis is provided in ~\ref{subsec:error}.

\begin{table}[htp!]
\scriptsize
\renewcommand{\arraystretch}{1.0}
\begin{tabularx}{\linewidth}{X}
\toprule
\textbf{Question:} Which month was John in Italy? \\ \midrule
\textbf{Gold Answer:} December, 2023 \\ \midrule
\multicolumn{1}{c}{\textsc{\textbf{DyCP}}} \\ \midrule
\textbf{Response:} John visited Italy in December 2023. \\ \midrule
\textbf{Evaluation:} The Bot accurately answers the User Question by correctly identifying that John visited Italy in December 2023. This information is supported by the conversation history, specifically the exchange on January 2, 2024, where John mentions his trip to Italy last month, which refers to December 2023. The Bot Response aligns with the Gold Response, affirming its correctness. [[rating]] 100 \\ \midrule
\multicolumn{1}{c}{\emph{\textbf{Full Context}}} \\ \midrule
\textbf{Response:} From our prior chats, your trip to Italy was in November 2023. We spoke about it on December 26th, 2023. \\ \midrule
\textbf{Evaluation:} The user asks for the month when John was in Italy, based on the conversation history. The Gold Response correctly states it as December 2023. The Bot incorrectly states that the trip was in November 2023. In the conversation history dated January 2, 2024, at 5:26 pm, John's trip to Italy in December 2023 is clearly mentioned: "By the way, have you been to Italy? I had a blast there last month." The Bot Response does not match this information and is therefore inaccurate. The inaccuracy is significant enough to merit a lower rating. [[rating]] 30 \\ \bottomrule
\end{tabularx}
\caption{An example where the \emph{Full Context} setting hallucinates.}
\vspace{-0.3cm}
\label{fig:full_failure_1}
\end{table}

\subsection{Recall or Precision?}
\label{subsec:recall_precision}
\begin{figure}[h!]
    \centering
    \includegraphics[width=1.0\linewidth]{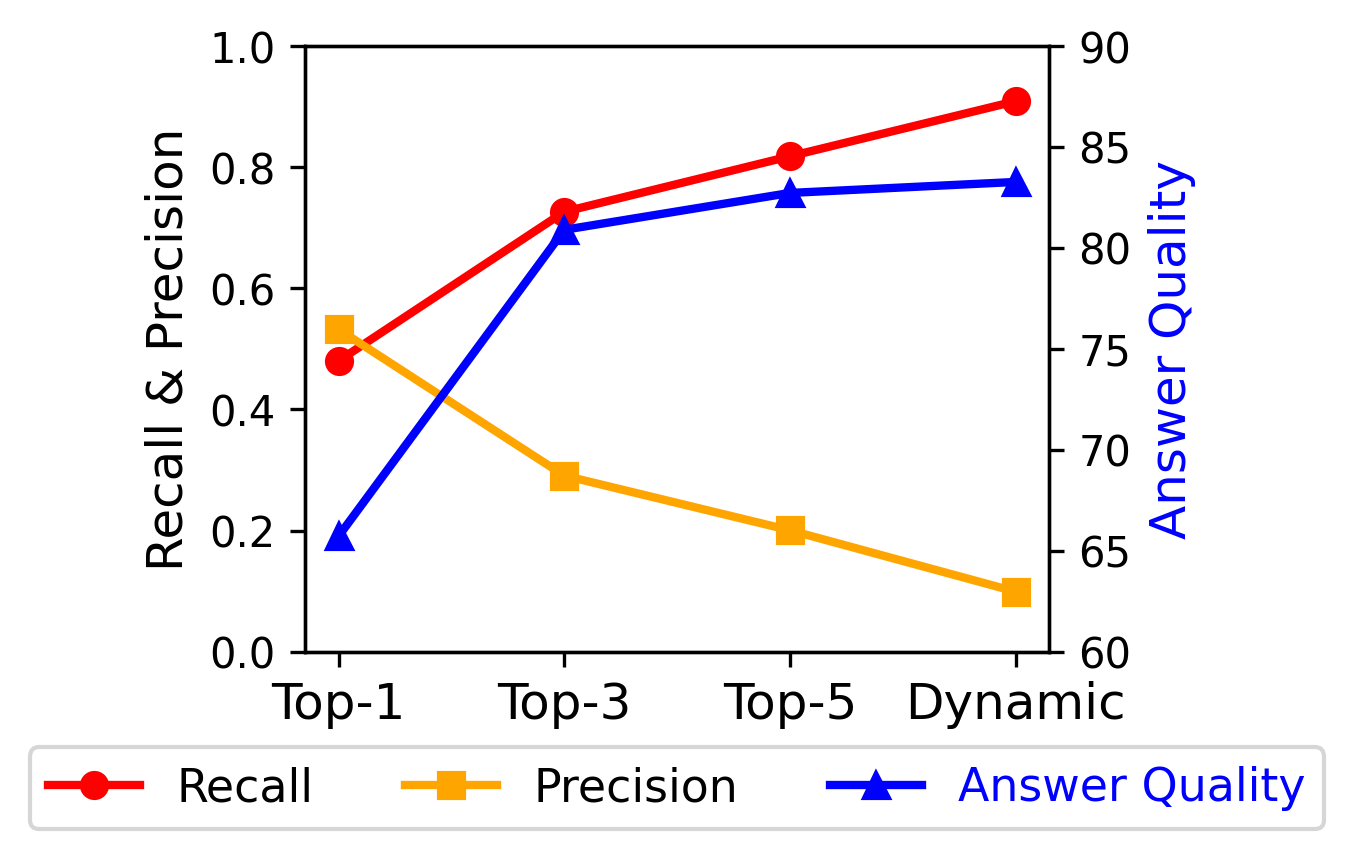}
    \caption{Relationship between retrieval recall, precision, and final answer quality in \textsc{\textbf{DyCP}}.}
    \label{fig:analysis_recall_precision}
\vspace{-0.2cm}
\end{figure}

To understand what makes pruning effective, we examine how different retrieval configurations influence the trade-off between recall and precision, and how this trade-off, in turn, affects both answer quality and latency. Figure~\ref{fig:analysis_recall_precision} shows the relationship between retrieval recall, precision, and final answer quality across different retrieval configurations in \textsc{DyCP}. 

\begin{figure}[h!]
    \centering
    \includegraphics[width=0.85\linewidth]{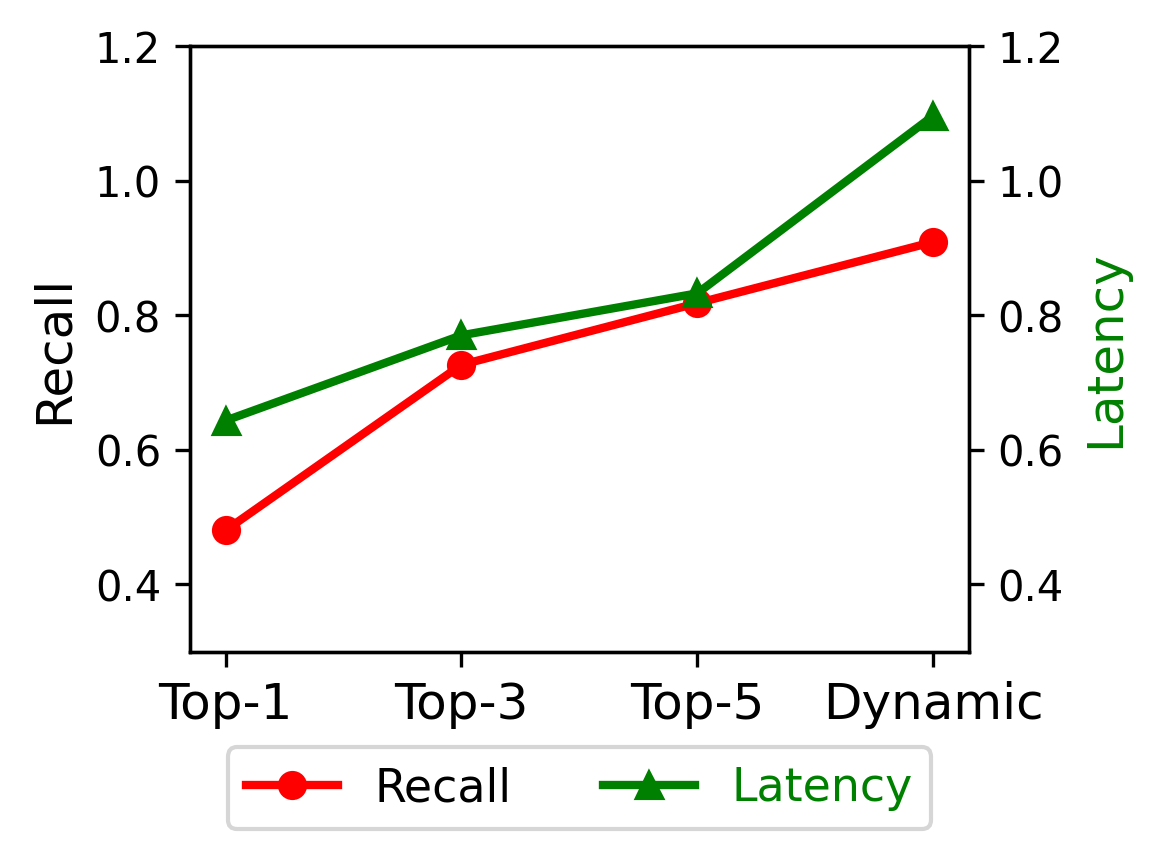}
    \caption{Relationship between retrieval recall and response latency.}
    \label{fig:analysis_recall_latency}
\vspace{-0.2cm}
\end{figure}

We observe a strong positive correlation between recall and answer quality: as recall increases, from Top-1 to Dynamic selection, answer quality steadily improves. This trend suggests that ensuring the inclusion of relevant content (high recall), even at the cost of introducing some mildly irrelevant information (lower precision), is critical for downstream answer generation. However, further increases in recall yield diminishing improvements, indicating that retrieving additional segments beyond a certain point offers limited value. Furthermore, because higher recall typically increases response latency (Figure~\ref{fig:analysis_recall_latency}), the goal is not simply to maximize recall, but to identify the point at which additional context no longer contributes meaningful gains.

\vspace{-0.3cm}
 
\noindent These observations also help explain why the concept underlying \textsc{DyCP}'s dynamic span selection is effective. Unlike fixed top-k approaches, \textsc{DyCP} expands the retrieved span only when the relevance dynamics justify it—adding segments only when sustained positive gain is detected. When relevance drops, the method stops expanding the span, preventing unnecessary context accumulation. This form of adaptive control maintains high recall where it matters while avoiding the latency penalty associated with uniformly larger contexts. In this sense, the concept of \textsc{DyCP} provides a principled mechanism for balancing recall and efficiency in long-form, multi-topic dialogue.

\subsection{Ablation on Continuity-Preserving Turns}
\label{subsec:ablation}
To evaluate the contribution of low-relevance but continuity-preserving turns in \textsc{DyCP}, we conduct an ablation study that progressively removes the least-relevant turns from each selected segment while measuring the resulting changes in answer quality and latency. These removed turns correspond to low but positive relevance scores—turns that are not directly needed for answering the current user query but help maintain local continuity and discourse flow within the segment. When removal would leave a segment empty, we retain the single most relevant turn to ensure that each segment preserves at least one anchor point.

\begin{figure}[h!]
    \centering
    \includegraphics[width=0.85\linewidth]{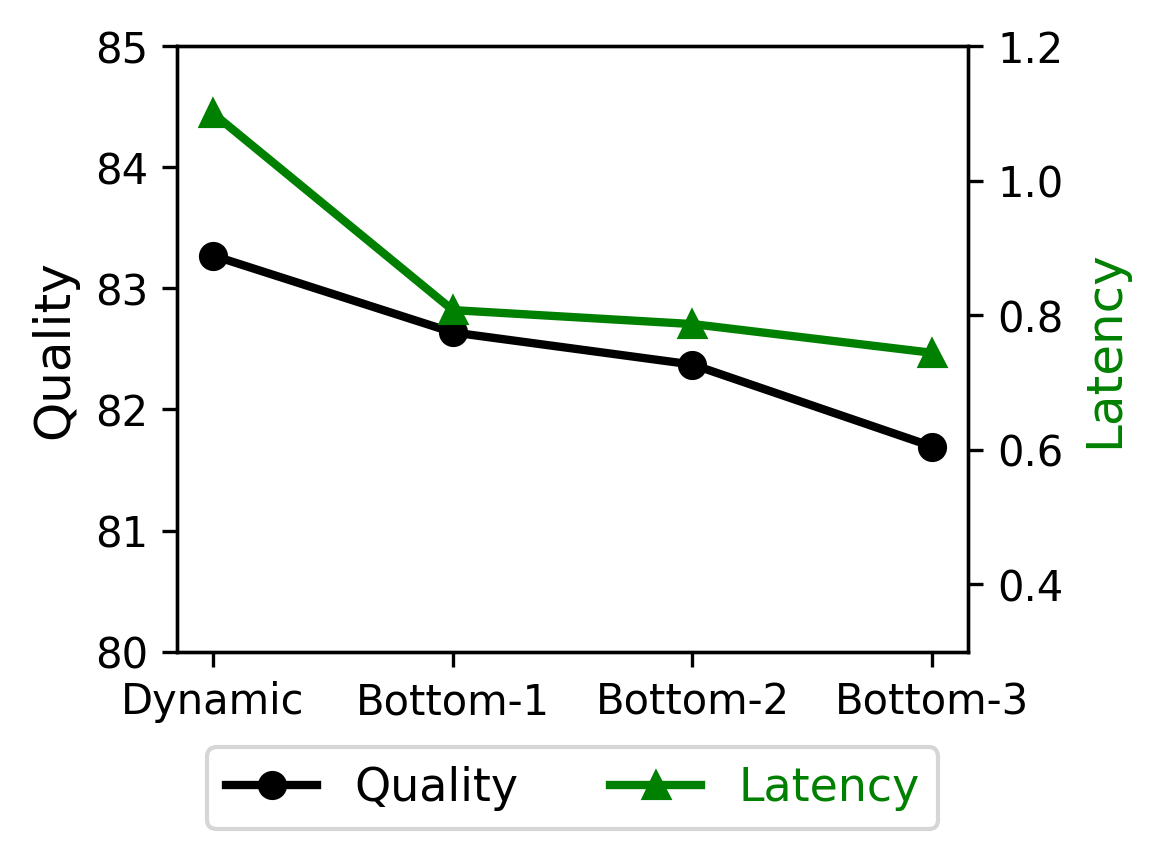}
    \caption{Ablation on continuity-preserving turns. We progressively remove the least-relevant turns within each \textsc{DyCP}-selected segment (Bottom-1/2/3) and compare the resulting answer quality (left axis) and response latency (right axis) against the original Dynamic setting. When removal would leave a segment empty, we retain the single most-relevant turn so that each segment preserves at least one anchor point.}
    \label{fig:ablation_turn}
\vspace{-0.3cm}
\end{figure}

Figure~\ref{fig:ablation_turn} shows that removing these turns leads to small but consistent drops in answer quality, indicating that even weakly relevant turns contribute useful contextual signals that support downstream generation. Latency decreases slightly as more turns are removed, as expected from the shorter input. Overall, these turns appear to help maintain a coherent local context, and the mild quality drop suggests that the pruning level can be tuned to balance efficiency and quality under different deployment budgets.

%% file: latex/section/discussion.tex
We raise the question: \textit{"Do LLMs meaningfully process far less context than their actual input token limits?"} While models like GPT‑4o support up to 128k tokens, our results indicate that they struggle to fully utilize even 25k tokens, especially when relevant information appears earlier in the input. This observation is consistent with prior findings such as \citet{an2024doeseffectivecontextlength}, who show that effective context lengths often fall short, typically not exceeding half of their training lengths. Interestingly, GPT‑4.1, with its 1M token limit, handles the same 25k-token dialogues much more reliably. This contrast suggests that token limits may reflect throughput capacity rather than effective working memory, and that in practice, models may attend to only a smaller, attention-weighted subset of the input. Several factors may contribute to this discrepancy, including the challenge of learning long-range dependencies during training, position-sensitive encoding, and attention saturation. Thus, while token limits continue to grow, effective long-context understanding may remain limited, both due to architectural bottlenecks and practical concerns such as inference cost.

%% file: latex/section/conclusion.tex
In this work, we  present \textsc{DyCP}, a dynamic context pruning method for long-form dialogue that selects query-relevant segments in real time, without pre-segmentation or incurring additional LLM calls. Across three benchmarks, \textsc{DyCP} consistently shows strong answer quality while significantly reducing response latency. Our findings show that, high recall is more beneficial than precision for retrieval, and preserving the sequential nature of dialogue plays a key role in response generation. We further show that, despite ongoing improvements in long-context LLMs, context management remains important for effective control of latency and inference cost. We offer a simple yet effective segment-level context management strategy that is dynamic, scalable, and LLM-agnostic, making it well-suited for real-world deployment. The scope and limitations of our work are discussed in Section~\ref{sec:limitation}.

%% file: latex/section/limitation.tex
\label{sec:limitation}

Our goal is to study a lightweight, retriever-based strategy for reducing the effective input length of long-form dialogue history at inference time. While \textsc{DyCP} consistently reduces the number of prepended context tokens and achieves low first-token latency in our primary API-based setting, several deployment factors can change the speed--cost trade-off in practice.

\paragraph{KV cache reuse and what latency we measure.}
Our latency metric is \emph{first token latency} measured via streaming API calls (Section~\ref{subsubsec:exp_quality_latency}). This reflects a common stateless serving pattern where each request is processed independently and developers cannot assume that key--value (KV) attention caches from previous turns are reused across calls. In contrast, in stateful local serving, a \emph{Full Context} strategy can be competitive or even faster because most of the history has already been prefetched and cached; in that case, modifying the input (as \textsc{DyCP} does) may invalidate KV reuse and increase the prefill cost, potentially raising the latency. Therefore, our speedups should be interpreted as evidence for settings where KV reuse is unavailable, unreliable, or provider-managed in ways that are not exposed to the developer. A practical next step is to make pruning \emph{cache-aware}, e.g., using \textsc{DyCP} when cache reuse is unlikely (e.g., long inter-turn gaps, non-persistent session routing) and falling back to a stable-prefix strategy when cache reuse is strong.

\paragraph{Provider-side prompt caching and billing effects.}
Modern API providers may apply \emph{prompt caching} (prefix caching) where repeated prompt prefixes are served more cheaply and/or with lower latency. This mechanism typically requires \emph{exact prefix matches} and operates under provider-defined cache lifetimes. Because \textsc{DyCP} dynamically selects different history spans per query, it can reduce cache-hit rates compared to strategies that keep a more stable prefix across turns. Under stable prompt caching, a \emph{Full Context} baseline may become cheaper, and the cost advantage of dynamic pruning can narrow or even reverse. Importantly, these caching policies are external to our method and can vary by provider, model, and request routing, so the realized cost--latency benefits depend on conditions that are not fully controllable by developers. We therefore emphasize that \textsc{DyCP} is a reliable way to reduce effective input length and improves response latency in stateless settings, while acknowledging that provider caching can change the monetary and latency calculus in production.

\paragraph{Overhead and robustness of pruning.}
\textsc{DyCP} does not introduce additional \emph{LLM} calls for segmentation or refinement, but it does require (i) storing turn embeddings and (ii) computing a query embedding and relevance scores at each turn. In our setup, this overhead is minimal relative to LLM inference, but it is not zero and it adds system complexity (retriever serving and vector storage). In addition, like other retriever-based approaches, \textsc{DyCP} can fail when semantic similarity does not surface the needed evidence. These failure modes also explain cases where \textsc{DyCP} underperforms the \emph{Full Context} baseline. Improving the retriever (or applying domain adaptation) is complementary to our contribution and directly improves pruning reliability.

\paragraph{Evaluation and generality.}
We use GPT4Score for scalable evaluation and include a small human study as a sanity check. The human evaluation is limited in scale and is intended as a qualitative complement to the automatic evaluation.

\paragraph{When \textsc{DyCP} is most useful.}
Taken together, \textsc{DyCP} is best viewed as a practical context-selection primitive for (a) stateless settings with limited KV reuse and provider-side caching, and (b) deployments where reducing input length is valuable for predictable latency and controllable budgeting.

%% file: latex/section/appendix.tex
\subsection{Example}
\label{subsec:example}
Table~\ref{fig:full_failure_2} illustrates a failure case of \emph{Full Context}, where the model incorrectly claims that no relevant information is available.

\begin{table}[h!]
\scriptsize
\renewcommand{\arraystretch}{1.0}
\begin{tabularx}{\linewidth}{X}
\toprule
\textbf{Question:} What are Joanna's plans for her finished screenplay in January 2022? \\ \midrule
\textbf{Gold Answer:} Submit it to film festivals and get producers and directors to check it out \\ \midrule
\multicolumn{1}{c}{\textsc{\textbf{DyCP}}} \\ \midrule
\textbf{Response:} Based on our conversation, in January 2022, Joanna planned to submit her finished screenplay to various film festivals in the hope that producers and directors would take notice of it. She was eager to receive feedback and explore new opportunities for her work to potentially make it to the big screen. \\ \midrule
\textbf{Evaluation:} The Bot Response accurately answers the User Question, closely matching the Gold Response. Both responses correctly identify Joanna's plans for her finished screenplay in January 2022: to submit it to film festivals in hopes that producers and directors would notice her work. (...) The bot provides a complete and relevant answer to the user question. [[rating]] 100 \\ \midrule
\multicolumn{1}{c}{\emph{\textbf{Full Context}}} \\ \midrule
\textbf{Response:} I'm sorry, but I don't have any information on Joanna's plans for her screenplay from January 2022. \\ \midrule
\textbf{Evaluation:} The Bot Response inaccurately states that there is no information available on Joanna's plans for her screenplay from January 2022, when, in fact, the conversation history contains this information. (...) Therefore, the Bot Response fails to accurately answer the User Question based on the provided conversation history. The lack of accurate recognition of the data contained in the conversation history warrants a low score for the bot's response. [[rating]] 10 \\ \bottomrule
\end{tabularx}
\caption{An example where the \emph{Full Context} setting incorrectly finds no relevant information.}
\vspace{-0.3cm}
\label{fig:full_failure_2}
\end{table}

\subsection{Error Analysis}
\label{subsec:error}

\begin{table}[h!]
\scriptsize
\renewcommand{\arraystretch}{1.0}
\begin{tabularx}{\linewidth}{X}
\toprule
\textbf{Question:} What causes does John feel passionate about supporting? \\ \midrule
\textbf{Gold Answer:} Veterans, schools, infrastructure \\ \midrule
\multicolumn{1}{c}{\textsc{\textbf{DyCP}}} \\ \midrule
\textbf{Response:} From our conversation, John seems passionate about supporting education and mentorship for underserved communities, raising awareness and funds for issues like homelessness and domestic abuse, and engaging in activities that support veterans' rights and the local fire station. John is driven by a desire to make a positive impact in the community through various forms of volunteering. \\ \midrule
\textbf{Evaluation:} The Bot Response provides a detailed answer by highlighting various causes that John feels passionate about, but it slightly embellishes the initial inquiry by listing numerous activities. The User Question asks specifically about "causes" John is passionate about supporting. The Gold Response provides a succinct list of main causes: Veterans, schools, and infrastructure. The Bot Response mentions some related causes but also includes additional activities like homelessness, domestic abuse, and supporting the fire station, which are mentioned in the broader context of the conversation but are not directly cited in the Gold Response as main causes. ($\ldots$) Therefore, the Bot Response is somewhat helpful but doesn\'t precisely match the concise focus intended by the Question and the Gold Response. [[rating]] 75 \\ \midrule
\multicolumn{1}{c}{\emph{\textbf{Full Context}}} \\ \midrule
\textbf{Response:} Throughout our conversations, John has expressed a strong passion for supporting various causes, including military veterans' rights, education reform, infrastructure development, and community service, such as participating in firefighting and raising funds through charity events. He believes in making a positive impact in our community and helping those in need. \\ \midrule
\textbf{Evaluation:} The Bot Response accurately answers the User Question by listing the causes John is passionate about: military veterans' rights, education reform, infrastructure development, and community service. This aligns with the conversation history, where John frequently discusses his involvement and interest in these areas. Additionally, the Bot Response expands on the Gold Response by adding community service, which is consistent with John's involvement in initiatives like firefighting and charity events. Therefore, the Bot Response is comprehensive and correct. [[rating]] 95 \\ \bottomrule
\end{tabularx}
\caption{An example where the \textsc{DyCP} generates an expanded response.}
\vspace{-0.3cm}
\label{fig:ours_failure}
\end{table}

Among the 23\% of cases where \textsc{DyCP} lose (Figure~\ref{fig:win_lose}), 66\% of these errors are attributable to imperfect retrievals, highlighting the importance of retrieval quality. An additional 34\% arise when \textsc{DyCP} generates an essentially correct answer that is rated marginally lower than the \emph{full context} response by GPTScore. To better understand the retrieval-related failures, we categorize them into three types: Missing Critical Turns (8\%), Partial Recall (16\%), and Retrieval Noise (42\%).

\vspace{0.2cm}
\noindent \textbf{Missing Critical Turns} $\:$ The retrieved context omits essential dialogue turns that are required to answer the question. This category includes approximately 5\% of cases where the retrieval fails to hit the gold context, i.e., failures in Hit (H) as reported in Table~\ref{tab:result_locomo_retrieval_by_method}.

\vspace{0.2cm}
\noindent \textbf{Partial Recall} $\:$ While the retrieved segments contain relevant information, they lack full coverage necessary for a complete answer. This often occurs when the answer depends on multiple segments scattered across the dialogue, and not all of them are successfully retrieved. In some cases, the issue is not due to retrieval failure, but rather stems from the subjective nature of the question, where referring to segments other than the gold context could still yield a reasonable answer.

\vspace{0.3cm}
\noindent \textbf{Retrieval Noise} $\:$ Although the necessary information is included in the retrieved context, the model fails to handle it effectively due to surrounding content. These cases frequently resemble failures observed in the \emph{full context} setting, where excessive input dilutes the signal. As in Table~\ref{fig:ours_failure}, the model tends to generate factually correct but verbose answers that obscure the main point.